\documentclass{article} 
\usepackage[preprint]{colm2026_conference}

\usepackage{microtype}
\usepackage{hyperref}
\usepackage{url}
\usepackage{booktabs}

\usepackage{colortbl}
\usepackage{graphicx}
\usepackage{wrapfig}
\usepackage{hyperref}
\hypersetup{colorlinks,linkcolor={red},citecolor={blue}, urlcolor={orange}}  
\definecolor{COLOR_MEAN}{HTML}{f0f0f0}
\definecolor{LIGHT_BLUE}{HTML}{e6f1fe}
\definecolor{LIGHT_RED}{HTML}{fceeee}
\definecolor{LIGHT_YELLOW}{HTML}{f1f58a}
\definecolor{LIGHT_GREEN}{HTML}{eaffea}
\definecolor{LIGHT_BROWN}{HTML}{f5e6d3}
\usepackage{multirow} 
\usepackage{makecell} 
\usepackage{caption}
\usepackage{tcolorbox}
\usepackage[subrefformat=parens]{subcaption}
\tcbuselibrary{breakable}

\usepackage[utf8]{inputenc} 
\usepackage[T1]{fontenc}    
\usepackage{hyperref}       
\usepackage{url}            
\usepackage{booktabs}       
\usepackage{amsfonts}       
\usepackage{nicefrac}       
\usepackage{microtype}      
\usepackage{xcolor}         
\usepackage{colortbl}
\usepackage{arydshln}
\usepackage{amsmath}
\usepackage{cleveref}
\usepackage{markdown}
\usepackage{tcolorbox}
\tcbuselibrary{breakable}
\usepackage{pdfpages}
\PassOptionsToPackage{numbers, compress}{natbib}
\usepackage{colortbl}
\usepackage{graphicx}
\usepackage{wrapfig}
\usepackage{hyperref}
\hypersetup{colorlinks,linkcolor={red},citecolor={blue}, urlcolor={orange}}  
\definecolor{COLOR_MEAN}{HTML}{f0f0f0}
\definecolor{LIGHT_BLUE}{HTML}{e6f1fe}
\definecolor{LIGHT_RED}{HTML}{fceeee}
\definecolor{LIGHT_YELLOW}{HTML}{f1f58a}
\definecolor{LIGHT_GREEN}{HTML}{eaffea}
\definecolor{LIGHT_BROWN}{HTML}{f5e6d3}
\usepackage{multirow} 
\usepackage{makecell} 
\usepackage{caption}
\usepackage{tcolorbox}
\usepackage[subrefformat=parens]{subcaption}
\tcbuselibrary{breakable}
\usepackage{tabularx}
\usepackage{ragged2e}

\usepackage{lineno}

\definecolor{darkblue}{rgb}{0, 0, 0.5}
\hypersetup{colorlinks=true, citecolor=darkblue, linkcolor=darkblue, urlcolor=darkblue}

\usepackage[whole]{bxcjkjatype}

\newcommand{\eg}{\textit{e.g.,}}
\newcommand{\ie}{\textit{i.e.,}}

\title{Paper Reconstruction Evaluation: Evaluating Presentation and Hallucination in AI-written Papers}

\author{%
Atsuyuki Miyai \quad
Mashiro Toyooka\thanks{Equal contribution} \quad
Zaiying Zhao\footnotemark[1] \quad
Kenta Watanabe\footnotemark[1] \\
\textbf{Toshihiko Yamasaki} \quad
\textbf{Kiyoharu Aizawa} \quad
\\
The University of Tokyo \\
\vspace{-0.8em} \\
\url{https://agent4science-utokyo.github.io/PaperRecon_HP}
}

\begin{document}

\ifcolmsubmission
\linenumbers
\fi

\maketitle

\begin{figure*}[h]
\vspace{-25pt}
\centering
    \includegraphics[width=0.99\linewidth]{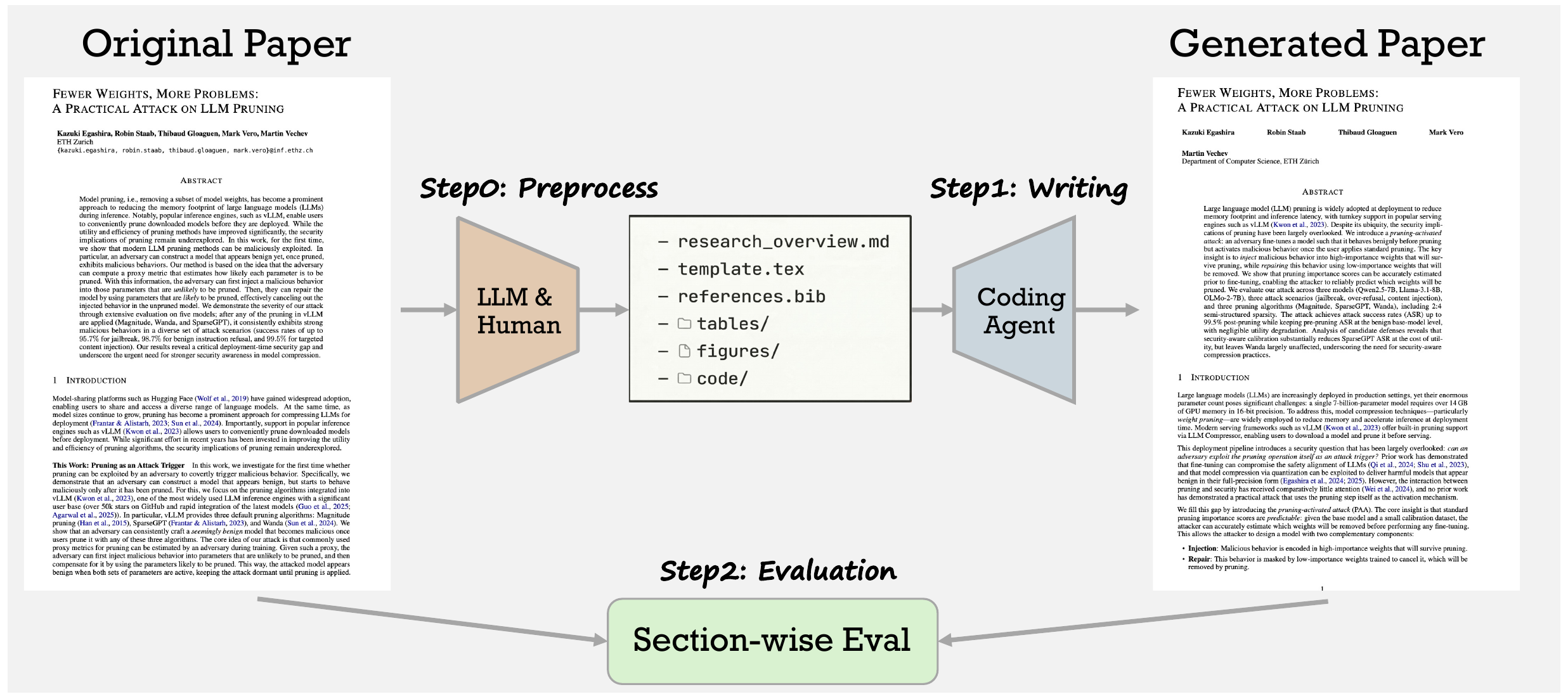}\\
   \caption{\textbf{Paper Reconstruction Evaluation}. Given minimal resources derived from a original paper, a coding agent reconstructs the full paper. The generated paper is then compared with the original paper to evaluate writing performance along two complementary axes: presentation quality and hallucination.}
    \label{fig:fig_teaser}
\end{figure*}

\begin{abstract}
    This paper introduces the first systematic evaluation framework for quantifying the quality and risks of papers written by modern coding agents.
    While AI-driven paper writing has become a growing concern, rigorous evaluation of the quality and potential risks of AI-written papers remains limited, and a unified understanding of their reliability is still lacking.
    We introduce \textbf{Paper Reconstruction Evaluation} (PaperRecon), an evaluation framework in which an overview (overview.md) is created from an existing paper, after which an agent generates a full paper based on the overview and minimal additional resources, and the result is subsequently compared against the original paper. PaperRecon disentangles the evaluation of the AI-written papers into two orthogonal dimensions, \textbf{Presentation} and \textbf{Hallucination}, where Presentation is evaluated using a rubric and Hallucination is assessed via agentic evaluation grounded in the original paper source.
    For evaluation, we introduce \textbf{PaperWrite-Bench}, a benchmark of 51 papers from top-tier venues across diverse domains published after 2025. Our experiments reveal a clear trade-off: while both ClaudeCode and Codex improve with model advances, ClaudeCode achieves higher presentation quality at the cost of more than 10 hallucinations per paper on average, whereas Codex produces fewer hallucinations but lower presentation quality. This work takes a first step toward establishing evaluation frameworks for AI-driven paper writing and improving the understanding of its risks within the research community.
\end{abstract}
\section{Introduction}
As recent AI tools, exemplified by coding agents, continue to advance, it is increasingly important to rigorously evaluate how they automate the research process and the potential risks they introduce, in order to ensure sustainable AI-driven scientific progress~\citep{miyai2025jr}. In particular, recent incidents involve submissions of AI-written papers to academic venues~\citep{liang2025quantifying, iclr2026_llm_response,  neubig2025tweet}. 
These submissions are likely to continue increasing alongside the rapid growth of AI Scientists in recent years~\citep{zochi2025, weng2025deepscientist, miyai2025jr}.
To advance scientific progress while preserving academic integrity, it is essential for the research community to continuously monitor the progress and risks of AI-driven writing based on rigorous and reliable evaluation.

Evaluating the paper writing capability of AI agents is inherently challenging and has not been sufficiently addressed in prior work. Existing approaches have explored using AI reviewers to assess paper quality~\citep{liu2024towards, yamada2025ai, weng2024cycleresearcher, zhu2025deepreview}. However, these methods are inadequate, as they tend to assign higher scores to papers with more severe fabrications~\citep{jiang2025badscientist, miyai2025jr}. 
While hallucinations in AI-written papers have been recognized, prior work has been limited to surface-level issues such as citation errors~\citep{walters2023fabrication, ansari2026compound, sakai2026hallucitation} or individual hallucination cases~\citep{yamada2025ai, miyai2025jr}, without enabling systematic evaluation.

We propose \textbf{Paper Reconstruction Evaluation} (PaperRecon), the first evaluation framework for measuring the paper writing capabilities of AI agents. The overview of PaperRecon is shown in \Cref{fig:fig_teaser}.
PaperRecon first compresses an existing paper into a structured summary, \texttt{research\_overview.md}, which retains only essential information. Given this compressed representation, along with other minimal resources (tables, figures, bib file), agents are tasked with reconstructing the original paper.
This workflow of generating a paper from minimal inputs is equivalent to isolating the writing component of existing AI Scientist systems~\citep{yamada2025ai, weng2025deepscientist, miyai2025jr}. If agents can reliably reconstruct high-fidelity papers, this provides a strong signal of their writing capabilities.

The key strength of PaperRecon as an evaluation framework lies in its ability to enable precise assessment through direct comparison with the original paper. In particular, it decomposes the notion of writing quality into two orthogonal dimensions: \textbf{Presentation} and \textbf{Hallucination}.
Presentation is evaluated using rubric evaluation~\citep{fan2024sedareval, phung2023generative, arora2025healthbench}, assessing how faithfully the key elements of the original paper are preserved in the reconstructed paper. Hallucinations are assessed via agentic evaluation grounded in the original paper source, enabling fine-grained detection of factual inconsistencies. This design allows PaperRecon to jointly evaluate both presentation quality and factual correctness in a unified and reliable manner.

For PaperRecon, we introduce \textbf{PaperWrite-Bench}, a benchmark constructed from 51 papers published after 2025. PaperWrite-Bench consists of papers from a diverse set of top-tier venues, including NeurIPS, ICLR, CVPR, ICCV, ACL, and ACMMM, providing broad coverage across research domains. This benchmark enables systematic and comprehensive evaluation of modern writing agents across realistic and diverse fields.

We evaluate recent powerful and widely used agents, including Claude Code~\citep{anthropic_claude_code}, Claude Code Agent Teams~\citep{anthropic_claude_code_teams}, and Codex~\citep{openai2025codex}, across a range of underlying models, from Claude Sonnet 4~\citep{anthropic2025claude4} and Claude Sonnet 4.6~\citep{anthropic2026claude46} to GPT-5~\citep{gpt5} and GPT-5.4~\citep{openai_gpt54_2026}. Our experiments yield the following key findings:

\begin{enumerate}
\item \textbf{Claude Code achieves higher presentation quality than Codex.} Claude Code better captures the key elements required for scientific writing across sections.
\item \textbf{Codex produces fewer hallucinations than Claude Code.} While Claude Code exhibits more than 10 hallucinations per paper on average, Codex limits this to around 3.
\item \textbf{Writing capability improves with model advances.} This also suggests that Paper Reconstruction Evaluation serves as a reliable metric for tracking progress in writing ability.
\end{enumerate}

Our work makes the following contributions to the research community:
\begin{itemize}
\item \textbf{Paper Reconstruction Evaluation (PaperRecon):}
We propose the first evaluation framework for scientific writing, Paper Reconstruction Evaluation, which measures the quality of paper reconstruction from compressed representations, along with a detailed evaluation protocol.

\item \textbf{PaperWrite-Bench:}  
We introduce PaperWrite-Bench, a benchmark constructed from recent papers across diverse research domains, enabling comprehensive evaluation of agents' ability to reconstruct papers from minimal information.

\item \textbf{Quantitative analysis of presentation and hallucination:}  
We provide a systematic evaluation of modern agents across both Presentation and Hallucination, quantifying how their capabilities evolve with model advances. Our results offer insights into the current state and trade-offs of AI-driven scientific writing.
\end{itemize}

\section{Related Work}
\textbf{AI-driven Research Automation.}
Recent progress in AI has accelerated efforts to automate various stages of the research process~\citep{si2024can, si2025ideation, asai2024openscholar, novikov2025alphaevolve, weng2024cycleresearcher, villaescusa2025denario, mitchener2025kosmos, ZHUANG2025103332, Lin2023ASPR, gottweis2025towards}, as well as end-to-end research pipelines~\citep{lu2024ai, zochi2025, tang2025ai, miyai2025jr}. However, recent work highlights the growing importance of carefully understanding the risks associated with increasingly capable AI scientists~\citep{miyai2025jr}. In particular,~\cite{miyai2025jr} reports risks across all stages of the research process, including idea generation, experimentation, writing, and review, and shows that the writing stage is especially prone to hallucinations such as inconsistencies with experimental results and fabricated content.
In this work, we aim to accurately measure the writing capabilities of modern agents through a novel evaluation framework, Paper Reconstruction Evaluation.

\textbf{Evaluation of AI-written Articles.}
Evaluation of AI-written articles has been studied in domains beyond scientific papers~\citep{yang2023doc, fitria2023artificial, zhong2026ai, shao2024assisting}. For example, prior work has long explored generating Wikipedia pages~\citep{banerjee-mitra-2015-wikikreator, minguillon2017semi, j.2018generating, fan-gardent-2022-generating, shao2024assisting}. In addition,~\cite{zhong2026ai} evaluates how well AI systems can write essays by using real GRE writing prompts. However, there are few papers to evaluate AI-driven scientific writing. This difference is likely due to the complexity of scientific writing, which requires justifying the significance of the problem, situating the work within the existing research landscape, designing sound and valid evaluation protocols, ensuring reproducibility and verifiability, and maintaining alignment between claims and supporting evidence. These requirements made earlier AI systems struggle with scientific writing and its evaluation, making it substantially more challenging than tasks such as Wikipedia or essay writing.

\textbf{Hallucination and Evaluation of AI-written Papers.}
It is widely recognized that papers written by AI systems often contain hallucinations. However, existing studies have been limited to surface-level issues such as citation errors~\citep{walters2023fabrication, ansari2026compound, sakai2026hallucitation} or individual hallucination cases~\citep{miyai2025jr}, without enabling systematic evaluation. This limitation largely stems from the absence of a well-defined protocol for assessing the substantive content of generated papers. As a result, existing evaluations of AI-written scientific papers have primarily relied on review-based assessment, focusing on whether a paper surpasses the acceptance threshold of academic venues~\citep{lu2024ai, yamada2025ai, weng2024cycleresearcher, zhu2025deepreview}. However, prior work has shown that AI reviewers often fail to detect hallucinations, leading to cases where papers with more severe hallucinations receive higher scores~\citep{jiang2025badscientist, miyai2025jr}.
While  AI-Researcher~\citep{tang2025ai} shares a similar perspective in generating new papers from source materials of existing research, it evaluates novelty, methodological validity, and empirical performance, making them fundamentally different from writing evaluation.
Therefore, to accurately understand the risks of AI-driven writing, it is essential to move beyond review-based evaluation and establish an evaluation protocol that directly assesses both presentation quality and hallucination.

\section{Paper Reconstruction Evaluation}
\subsection{Problem Definition}
Paper Reconstruction Evaluation (PaperRecon) is a framework for evaluating how accurately coding agents can reconstruct scientific papers.

From each original paper, we extract the following information and provide it to the agent:
(1) \texttt{Research Overview}: A Markdown file summarizing the motivation, method, and key experimental results of the paper.
(2) \texttt{Figure}: Figures from the original paper with simplified captions (\eg~fig\_method.jpg: Method Overview.).
(3) \texttt{Table}: LaTeX source code of tables from the original paper with simplified captions (\eg~table\_cb.tex: Main Result.).
(4) \texttt{References}: The bibliography file of the original paper, where each entry is augmented with its abstract.
(5) \texttt{Code}: The codebase associated with the original paper, if available.

Given these inputs, the agent is tasked with reconstructing the original paper. The generated paper is then compared against the original from multiple perspectives to evaluate writing capability.
Here, we reuse the original references instead of requiring agents to construct them from scratch, as accurate reference collection is itself a separate research problem. This design allows us to isolate and evaluate the core writing ability of the agent.

\subsection{Evaluation Protocol}
\begin{figure*}[t]
\centering
    \includegraphics[width=0.99\linewidth]{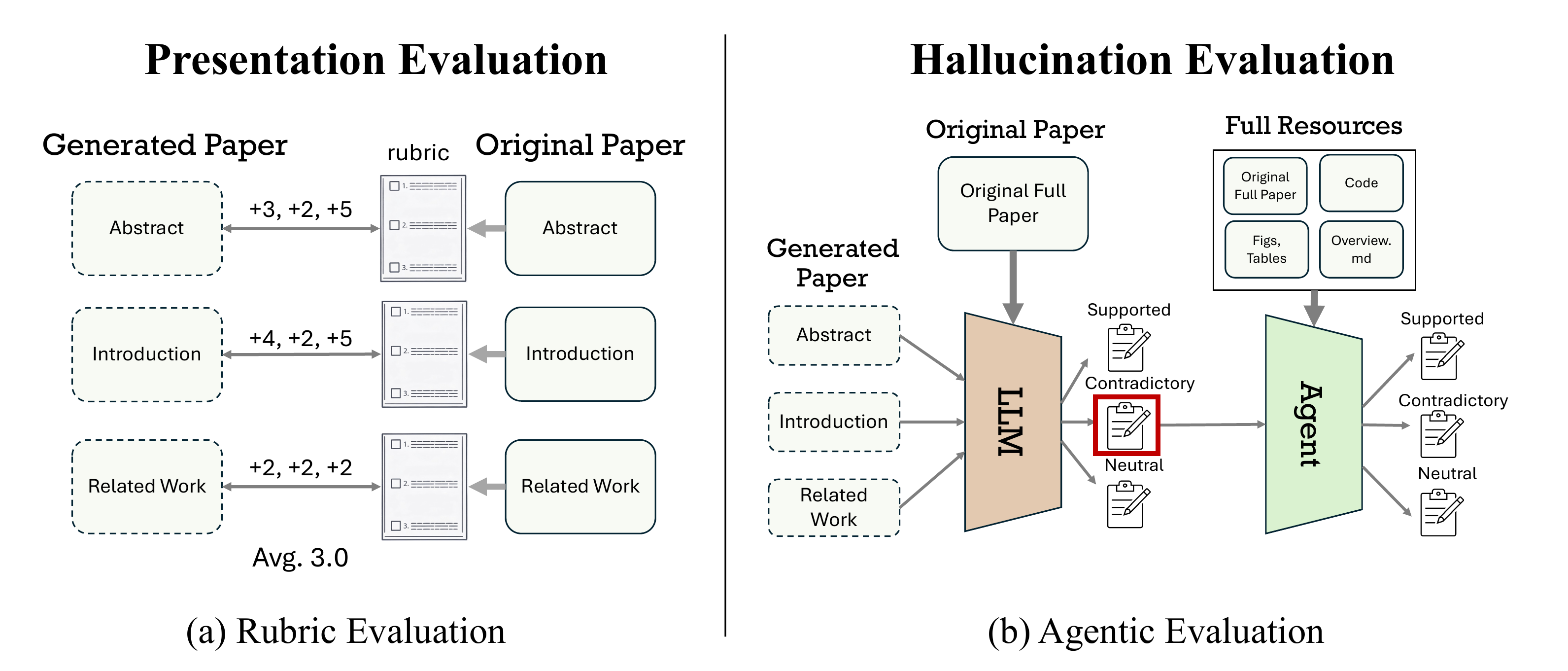}\\
   \caption{\textbf{Overview of PaperRecon evaluation pipeline.} Our evaluation compares generated papers against GT (original) papers along two complementary axes: Presentation evaluation with rubric and agentic hallucination evaluation}
    \label{fig:evaluation}
\end{figure*}

We evaluate generated papers by comparing them against ground-truth (GT) papers along three complementary axes: Presentation evaluation with rubric, agentic hallucination evaluation, and citation-level evaluation. \Cref{fig:evaluation} shows the overview of presentation and hallucination evaluation pipelines.
To compute these scores, we first perform section classification and matching, then evaluate each matched pair.

\subsubsection{Section Classification and Matching}
Papers organize sections differently, so we first extract all sections from both the GT and generated LaTeX files. Next, we map them into seven common categories: \textit{Abstract}, \textit{Introduction}, \textit{Method}, \textit{Benchmark Construction}, \textit{Experiment}, \textit{Related Work}, and \textit{Conclusion}. Classification first applies keyword-based rules (\eg~``preliminary'' $\to$ Method, ``ablation studies'' $\to$ Experiment); sections that cannot be classified by rules are passed to an LLM for classification based on section name and content. When multiple raw sections map to the same category, they are merged. This classification is performed once and shared across all subsequent evaluation steps.

\subsubsection{Rubric Evaluation}
To enable accurate and fine-grained evaluation, we adopt a rubric evaluation~\citep{fan2024sedareval, phung2023generative, arora2025healthbench}. 
We found in preliminary experiments that LLM-as-a-judge yielded low discriminative power in evaluation, whereas rubric evaluation enables more discriminative assessment.

For each GT paper, we pre-construct a rubric that specifies key elements expected in each section, along with their relative importance and descriptions. Each rubric element corresponds to a concrete and verifiable point (\eg~``Problem motivation: improve reasoning in vision-language models'' for the Abstract, or ``Evaluation covers both visual reasoning and general image understanding benchmarks'' for the Experiment section). These rubrics are initially generated from the GT paper using an LLM (\ie~GPT-5.4~\citep{openai_gpt54_2026}), and are subsequently reviewed and, when necessary, refined by the authors to ensure quality and correctness. 

For each section excluding the Conclusion section, we use an LLM judge with GPT-5.4 to evaluate how well the generated section covers each rubric element on a 1--5 scale:
\textit{5: the content is fully and accurately described with correct details};
\textit{4: the content is mostly described, with the core idea present but some details missing};
\textit{3: the content is partially described, with significant gaps or vagueness};
\textit{2: the content is barely mentioned, with only a superficial or indirect reference};
\textit{1: the content is completely absent from the generated section.}

In addition to text-based rubric elements, we incorporate figure- and table-level scores as part of the evaluation rubric.

\textbf{Figure Evaluation.}
We align GT and figures in the generated papers based on GT Tex file and assess their contextual appropriateness. If both the GT and generated papers reference a figure within the same section, a full context score of 5 is assigned. Otherwise, an LLM evaluates whether the figure is used in an appropriate context on a 1--5 scale. If the generated paper references the figure in any section (even if different from the GT), the LLM-derived score is used; if the figure is not referenced at all, it is assigned a score of 1.

\textbf{Table Evaluation.}
We extract tables from both GT and generated LaTeX files and perform matching using a hierarchical strategy: label matching, caption matching, and LLM-based matching. For each matched pair, an LLM evaluates numerical accuracy, structural alignment, and content consistency, producing a match score on a 1--5 scale. Tables present in the GT but missing from the generated paper are assigned a score of 1.

The final rubric score for each section is the average across all rubric elements (text, figure, and table items combined).
The average number of rubrics for each section is as follows: Abstract: 10.3, Introduction: 13.3, Related Work: 12.6, Method: 14.2, Benchmark Construction: 14.6, and Experiment: 14.3.

\subsubsection{Hallucination Analysis}

We identify factual errors via a two-stage, claim-level analysis.

\textbf{Stage 1: Claim Extraction.}
For each section (excluding the Conclusion), an LLM (\ie~GPT-5.4) extracts all concrete and verifiable claims from the generated content and classifies each claim into one of the following categories:

\textit{1. Supported}: directly stated in, or logically derivable from, the GT paper.

\textit{2. Neutral}: not present in the GT paper, but a reasonable general statement or supplementary detail that does not contradict it (note that absence from the GT does not imply contradiction).

\textit{3. Contradictory}: directly conflicts with specific information in the GT paper.

For claims classified as contradictory, we further assign a severity level: \textbf{major} (\eg~incorrect numerical values, fabricated results, or incorrect method descriptions) or \textbf{minor} (\eg~overly strong generalizations, imprecise wording).

\textbf{Stage 2: Verification.}
All claims labeled as contradictory across sections are aggregated and re-evaluated with a coding agent (\ie~Claude Code with Sonnet4.6). The agent is provided with the GT paper resources (including LaTeX source, codebase, figures, and tables) and re-examines each flagged claim, potentially revising false positives to \textit{supported} or \textit{neutral}. This two-stage design reduces false positives while limiting the computational cost to a single agent invocation.

Finally, we report the total counts of \textit{supported}, \textit{neutral}, \textit{major contradictory}, and \textit{minor contradictory} claims.

\subsubsection{Citation-Level Evaluation}
While the primary objective of PaperRecon is to assess the content of generated papers, we also evaluate citation usage by comparing citation keys between the GT and generated papers using an F1-based metric.

We extract all citation keys in both GT and generated LaTeX files, and compute precision, recall, and F1 score based on the overlap of citation key sets. We additionally detect \textit{hallucinated citations} (keys cited in the generated paper but absent from \texttt{references.bib}), \textit{missing citations} (keys in GT but not in the generated paper), and \textit{extra citations} (keys added by the prediction that are not in GT).

\subsubsection{Overall Metrics}
We report the following aggregate metrics:
(i) \textbf{Avg.\ Rubric Score}: Mean rubric score across all evaluated sections and elements (1--5 scale), including figure and table items.
(ii) \textbf{Hallucination Counts}: Total number of major contradictory claims detected across all sections after two-stage verification.
(iii) \textbf{Citation Scores}: Precision (valid cited / total cited), Recall (valid cited GT / total GT), F1 (harmonic mean), Hallucination (num of invalid citations).

\section{PaperWrite-Bench}
\subsection{Benchmark Overview}
In this section, we introduce PaperWrite-Bench, a benchmark designed for PaperRecon. PaperWrite-Bench consists of 51 papers manually curated by the authors, selected from a diverse set of top-tier conferences, including ACL 2025, EMNLP 2025, CVPR 2025, CVPR 2026, ICCV 2025, ICLR 2025, NeurIPS 2025, ICLR 2026, and ACMMM 2025, covering a wide range of domains such as computer vision, natural language processing, machine learning, and multimedia processing.
Among the 51 papers, 32 focus on proposing new methods, 12 introduce new benchmarks, and 7 combine both contributions. This diversity enables comprehensive evaluation of agent writing capabilities across different types of scientific papers.

Prior work has constructed similar benchmarks for the reproduction of the experimental results by collecting papers, such as Exp-Bench (56 papers)~\citep{kon2025exp} and PaperBench (20 papers)~\citep{starace2025paperbench}~\citep{kon2025exp, hu2025repro, starace2025paperbench}. However, these benchmarks are largely based on papers published around 2024 and do not reflect more recent developments.
Therefore, we curate PaperWrite-Bench from more recent sources to better reflect the current capabilities of modern agents.

\subsection{Benchmark Construction Pipeline}

\textbf{Construction of research\_overview.md.}
For each paper, we generate a \texttt{research\_overview.md} using GPT-5~\citep{gpt5}, which summarizes the key information required for reconstruction. To ensure quality, the authors manually verify that the overview contains sufficient information to faithfully reconstruct the original paper. On average, each file contains 463 words.

\textbf{Extraction of Tables, Figures, References, and Code.}
Using the arXiv source files, we extract tables and figures and store them in dedicated directories (\texttt{tables/} and \texttt{figures/}). Following prior work~\citep{liu2024towards}, we provide structured reference information to the agent via \texttt{table\_summary.txt} and \texttt{figure\_summary.txt}, which include file paths and brief descriptions (e.g., the first line of the original caption).
To ensure proper use of references, we augment each entry in the \texttt{.bib} file with abstracts retrieved via the Semantic Scholar API. In addition, we include the associated codebase, when available, to facilitate accurate reconstruction of the proposed methods.
When the README.md file in the code contained sections such as an abstract or introduction, those parts were manually removed.

\textbf{Template and Style File Construction.}
We extract the section structure of each original paper from the arXiv source and create a \texttt{template.tex}, along with the necessary style files.
The \texttt{template.tex} contains only the section structure of the original paper, preserving the same section headings while leaving the content empty. Agents are instructed to generate the paper following this predefined structure. This design is motivated by the observation that section organization varies significantly across papers, making direct comparison between generated and original papers difficult.
By fixing the section structure, we enable more accurate and consistent evaluation. Moreover, since defining the section structure is a simple preprocessing step, this pipeline remains practical, requiring only minimal human intervention before the writing task is carried out by the agent.
\section{Experiments}

\subsection{Experimental Setup}
\textbf{Agents.}
We evaluate three coding agents: Claude Code~\citep{anthropic_claude_code} (single-agent), Codex~\citep{openai2025codex} (single-agent), and Claude Agent Teams~\citep{anthropic_claude_code_teams} (multi-agent). 
Claude Code is evaluated with Sonnet 4~\citep{anthropic2025claude4} and Sonnet 4.6~\citep{anthropic2026claude46}, Codex with GPT-5~\citep{gpt5} and GPT-5.4~\citep{openai_gpt54_2026}, and Claude Agent Teams with Sonnet 4.6, resulting in five agent configurations in total.

\textbf{Writing Pipeline.}
The goal of this study is to understand how well the current agents can perform scientific writing under a simple setup and what risks arise in the process. Therefore, we adopt a deliberately simple writing pipeline. Our pipeline includes a compilation feedback loop, where LaTeX errors are returned to the agent for correction, as well as a page limit adjustment step following \citet{liu2024towards, yamada2025ai}.

\subsection{Main Results}

\begin{table*}[t]
\centering
\caption{\textbf{Presentation evaluation.} Rubric evaluation scores by model and section (1--5 scale). We observe that Claude Code achieves higher presentation quality than Codex.}
\vspace{-3mm}
\label{tab:rubric_summary}
\begin{tabular}{@{}llrrrrrrrr@{}}
\toprule
Agent & Model & Abs. & Intro. & Rel. & Meth. & Bench. & Exp. & Avg. \\
\midrule
Codex & GPT5  & 4.00 & 3.58 & 2.32 & 2.89 & 3.25 & 3.53 & 3.26 \\
Codex & GPT5.4  & 4.06 & 3.87 & 2.72 & 3.51 & 3.79 & 3.64 & 3.59 \\
\midrule
ClaudeCode & Sonnet4  & 4.10 & 3.88 & 2.48 & 3.23 & 3.63 & 3.66 & 3.49 \\
ClaudeCode & Sonnet4.6  & \textbf{4.37} & \textbf{4.12} & \textbf{3.08} & \textbf{3.69} & 3.84 & \textbf{4.00} & \textbf{3.86} \\
ClaudeCode-Teams & Sonnet4.6  & 4.28 & 4.05 & 3.07 & 3.62 & \textbf{3.99} & 3.97 & 3.82 \\
\bottomrule
\end{tabular}
\end{table*}

\begin{table*}[t]
\centering
\caption{\textbf{Hallucination evaluation.} Each score represents the average number of hallucinations per paper in each section. We observe that Codex produces fewer hallucinations than Claude Code.
}
\vspace{-3mm}
\label{tab:hal_summary}
\begin{tabular}{@{}llrrrrrrrrr@{}}
\toprule
Agent & Model & Abs. & Intro. & Rel. & Meth. &  Bench. & Exp. & Total \\
\midrule
Codex & GPT5 & 0.3 & 0.6 & 0.3 & 3.8 & 1.9 & 3.4 & 10.2 \\
Codex & GPT5.4 & \textbf{0.1} & \textbf{0.3} & \textbf{0.2} & \textbf{1.3} & \textbf{0.2} & \textbf{0.9} & \textbf{3.0} \\
\midrule
ClaudeCode & Sonnet4 & 0.2 & 0.5 & 0.5 & 5.4 & 0.8 & 4.7 & 12.0 \\
ClaudeCode & Sonnet4.6 & 0.2 & 0.8 & 0.6 & 4.7 & 0.5 & 3.6 & 10.4 \\
ClaudeCode-Teams & Sonnet4.6 & 0.3 & 0.6 & 0.8 & 3.9 & 0.5 & 3.8 & 9.8 \\
\bottomrule
\end{tabular}
\end{table*}

We report rubric-based presentation evaluation scores in \Cref{tab:rubric_summary}. For hallucination evaluation, we report the average number of hallucinations, defined as claims classified as \textit{major contradictory}, in \Cref{tab:hal_summary}.
\Cref{tab:citation_summary} shows the citation evaluation results.

\begin{table}[t]
\centering
\caption{Citation evaluation scores.}
\vspace{-3mm}
\label{tab:citation_summary}
\begin{tabular}{@{}llrrrrr@{}}
\toprule
Agent & Model & Prec. & Recall & F1 & Hal. \\
\midrule
Codex & GPT5  & \textbf{0.89} & 0.27 & 0.39 & \textbf{0.0} \\
Codex & GPT5.4  & 0.86 & 0.43 & 0.56 & \textbf{0.0} \\
\midrule
ClaudeCode & Sonnet4  & 0.75 & 0.24 & 0.34 & 3.5 \\
ClaudeCode & Sonnet4.6  & 0.83 & \textbf{0.58} & \textbf{0.67} & 0.2 \\
ClaudeCode-Teams & Sonnet4.6  & 0.84 & 0.56 & 0.66 & 0.2 \\
\bottomrule
\label{tab:citation}
\end{tabular}
\end{table}

We summarize the key findings below.

\textbf{[F1] Claude Code outperforms Codex in presentation quality.}
As shown in \Cref{tab:rubric_summary}, Claude Code consistently achieves higher presentation scores than Codex across all sections, indicating a stronger ability to capture and articulate key scientific points. However, the best-performing agent, Claude Code with Claude Sonnet 4.6, reaches a score of 3.86, suggesting that there remains substantial room for improvement.

\textbf{[F2] Claude Code exhibits substantially more Hallucinations, while Codex significantly reduces them.}
\Cref{tab:hal_summary} reports the average number of hallucinations per paper across sections. We observe a clear contrast: although Claude Code achieves higher presentation quality, it produces a large number of hallucinations, exceeding 10 per paper even with Claude Sonnet 4.6. In contrast, using GPT-5.4~\citep{openai_gpt54_2026} reduces hallucinations to around 3 per paper. 
These results reveal a clear trade-off between presentation quality and hallucination, highlighting the importance of evaluating both dimensions to accurately assess model performance.

\textbf{[F3] Codex produces fewer citation hallucinations than Claude Code.}
\Cref{tab:citation} reports the results on citation accuracy. 
Consistent with \textbf{[F2]}, while Claude achieves higher Citation F1 scores, Codex produces substantially fewer hallucinated citations. This again highlights a trade-off between citation coverage and factual reliability.

\textbf{[F4] Writing capability improves with model advances.} Our evaluation framework accurately captures performance gains from model improvements. We observe consistent increases in writing quality from Claude Sonnet 4~\citep{anthropic2025claude4} to Claude Sonnet 4.6~\citep{anthropic2026claude46}, as well as from GPT-5~\citep{gpt5} to GPT-5.4~\citep{openai_gpt54_2026}, demonstrating that PaperRecon effectively tracks progress in writing capability.

\subsection{Human Validation}

\textbf{Presentation Validation.}
To validate the reliability of our proposed evaluation framework, we conducted a human correlation analysis using 72 pairs of generated papers. These pairs were constructed from 12 source papers, each reconstructed by four agent configurations (\ie~Claude Code and Codex, each with two backbone LLMs), yielding six pairwise combinations per source. We recruited three human reviewers with experience as reviewers at top-tier conferences, each of whom provided pairwise judgments (win, tie, or lose) for the 24 paper pairs. We then measured the Kendall's $\tau_b$ correlation between these human judgments and the rankings induced by our rubric-based evaluation scores. 
Our results demonstrate a strong and highly significant correlation ($\tau_b = 0.578$, $p < 0.001$), indicating that the rubric-based evaluation aligns well with expert human judgment. We also observe discrepancies between rubric scores and human preferences, often driven by reviewers' subjective biases such as a preference for conciseness over detailed explanations. These findings suggest that our evaluation framework can provide consistent and high-quality scoring while mitigating subjective biases inherent in human evaluation.
 
\textbf{Hallucination Validation.}
For hallucination validation, we focus on measuring precision by verifying whether claims classified as \textit{major contradictory} are indeed incorrect. Exhaustively identifying all hallucinations is prohibitively labor-intensive; therefore, evaluating precision alone provides a reliable basis for model comparison. As this is a fact-checking task, the verification is conducted manually by the authors.
From the evaluation data, we extracted 97 instances labeled as \textit{major contradictory} in GPT-5, GPT-5.4 and Sonnet-4.6 papers and manually examined them. We find that 96\% of these correspond to genuine contradictions or fabrications. This result indicates that hallucinations detected by our method are highly likely to be true hallucinations.

\subsection{Analysis}
\begin{table*}[t]
\centering

\begin{minipage}[t]{0.48\textwidth}
\centering
\small
\caption{\textbf{Effect of research overview length.} Comparison of default vs.\ long research overview as input to the writing agent. Scores are averaged across 12 papers.}
\vspace{-3mm}
\label{tab:overview_length}
\begin{tabular}{l cc cc}
\toprule
& \multicolumn{2}{c}{\textbf{Rubric Eval}$\uparrow$} & \multicolumn{2}{c}{\textbf{Hallucination}$\downarrow$} \\
\cmidrule(lr){2-3} \cmidrule(lr){4-5}
\textbf{Overview} & Default & Long & Default & Long\\
\midrule
Sonnet4  & 3.49 & \textbf{3.64} & 8.8 & \textbf{5.8} \\
Sonnet4.6 & 3.83 & \textbf{4.17} & 9.8 & \textbf{2.3} \\
\bottomrule  
\end{tabular}
\end{minipage}
\hfill
\begin{minipage}[t]{0.48\textwidth}
\centering
\small
\caption{\textbf{Analysis by paper type.} Evaluation results grouped by conference type.}
\vspace{-3mm}
\label{tab:paper_type}
\begin{tabular}{l c cc}
\toprule
\textbf{Conf.} & \textbf{\# Papers} & \textbf{Rubric}$\uparrow$ &  \textbf{Hal.}$\downarrow$ \\
\midrule
ML & 21 & 3.58 & 8.3 \\
CV & 21 & 3.63 & 10.1 \\
MM & 5 & 3.47 & 10.7\\
NLP & 4 & \textbf{3.77} & \textbf{6.0} \\
\bottomrule
\end{tabular}
\end{minipage}

\end{table*}

\textbf{Effect of Research Overview Length.}
We investigate how the granularity of the research overview affects reconstruction quality by comparing default and long versions. The default overview provides a high-level summary (463 words on average), while the long overview includes more detailed descriptions of methodology and experimental design (1492 words on average).  Each example is shown in~\Cref{sec_append:overview}. \Cref{tab:overview_length} shows the results. Consistent with intuition, more detailed research overviews lead to higher presentation scores and fewer hallucinations. This also indicates that our evaluation metrics are able to accurately assess the quality of the papers.

\textbf{Performance by Conference Type.}
\Cref{tab:paper_type} shows the performance across different conference types. Although there are differences in the number of papers, we observe that NLP conferences achieve the highest performance.
To investigate the cause, we examined the original papers and found that NLP papers tend to focus more on findings-based research, with fewer complex mathematical formulations or methods compared to other fields.
Therefore, we consider that it is ultimately necessary to evaluate writing progress separately for each research field.
\section{Conclusion, Limitations and Future Work}
In this work, we introduce Paper Reconstruction Evaluation, the first systematic evaluation framework for AI-generated scientific papers. Along with PaperWrite-Bench, we conduct a comprehensive evaluation of the capabilities and risks of modern writing agents. We discuss the limitations of our approach and outline directions for future work below.

\textbf{Controlled Input Assumptions.}
Our framework provides structured resources, including figures, tables, and references, to the agent. This design reduces external dependencies such as retrieval and reference collection, and allows us to focus on evaluating core writing ability. Evaluating writing performance under more limited resources, including settings where models rely on external systems, is an important direction for future work.

\textbf{Limited Coverage of Diverse Writing Styles.}
Evaluating scientific papers is inherently challenging, as human writing is diverse and not fully captured by current LLMs. As a result, section-wise evaluation may not fully reflect overall quality. Developing more robust methods remains an important direction for future work.

\section*{Ethics Statement}
This work studies the capabilities and risks of AI-driven scientific writing through a controlled evaluation framework. While our approach enables systematic assessment of presentation quality and hallucinations, it also highlights the potential for advanced AI systems to generate plausible but misleading scientific content.

A key ethical concern is the misuse of such systems to produce fabricated or low-quality research papers that may bypass standard review processes. Our findings, particularly the trade-off between presentation quality and hallucination, underscore the importance of developing robust evaluation methods and safeguards against unreliable AI-generated content.

\section*{Authorship Statement}
\textbf{Atsuyuki Miyai} served as the project lead and director of the entire initiative, overseeing all aspects from the conception of PaperRecon to its execution and paper writing.

\textbf{Mashiro Toyooka} was primarily responsible for implementation, contributing to the development of the core codebase for agentic writing as well as the evaluation framework, in collaboration with Atsuyuki Miyai.

\textbf{Zaiying Zhao} implemented the figure evaluation system and contributed to the writing of the paper.

\textbf{Kenta Watanabe} contributed by carefully reviewing documents such as overview.md for completeness and proposing the Citation Score.

\textbf{Toshihiko Yamasaki} mainly served in a managerial role, providing valuable suggestions to the project on a regular basis.

\textbf{Kiyoharu Aizawa} provided regular and decisive guidance throughout the project, offering invaluable resource support that was critical to its successful execution.

\section*{Acknowledgment}
We would like to thank Qing Yu, Satoshi Kosugi, and Jeonghun Baek for reviewing our generated paper.
This work was partialy supported by JSPS KAKENHI 25H01164 and JST BOOST, Japan Grant Number JPMJBS2418.

\bibliography{colm2026_conference}
\bibliographystyle{colm2026_conference}

\clearpage
\newcommand\beginsupplement{%
        \setcounter{table}{0}
        \renewcommand{\thetable}{\Alph{table}}%
        \setcounter{figure}{0}
        \renewcommand{\thefigure}{\Alph{figure}}%
     }
\beginsupplement
\appendix
\section*{Appendix}
\section{PaperWrite-Bench}
\subsection{Statistics of PaperWrite-Bench Sources}
\Cref{tab:paper_distribution} presents statistics on the conferences included in PaperWrite-Bench Sources. As shown, ML and CV conferences constitute the majority, followed by multimedia and NLP conferences.

\begin{table}[h]
\centering
\caption{Distribution of papers by research area.}
\label{tab:area_distribution}
\begin{tabular}{llcc}
\toprule
Area & Conference & \# Papers & Subtotal \\
\midrule
\multirow{4}{*}{ML}
  & ICLR26    &  7 & \multirow{4}{*}{21} \\
  & NeurIPS25 &  6 & \\
  & ICLR25    &  4 & \\
  & ICML25    &  2 & \\
  & AAAI25    &  2 & \\
\midrule
\multirow{3}{*}{CV}
  & CVPR25  & 16 & \multirow{3}{*}{21} \\
  & ICCV25  &  3 & \\
  & CVPR26  &  2 & \\
\midrule
Multimedia & ACMMM25 & 5 & 5 \\
\midrule
\multirow{2}{*}{NLP}
  & ACL25   & 3 & \multirow{2}{*}{4} \\
  & NAACL25 & 1 & \\
\midrule
\multicolumn{2}{l}{Total} & 51 & 51 \\
\bottomrule
\label{tab:paper_distribution}
\end{tabular}
\end{table}

\begin{table*}[h]
\centering
\small
\caption{Representative rubric evaluation points per section (importance: high). Shown for a benchmark construction paper (EgoLife~\citep{yang2025egolife}, CVPR25) as a concrete example.}
\label{tab:rubric_example}
\begin{tabular}{@{}lp{12cm}@{}}
\toprule
Section & Evaluation Point \\
\midrule
Abstract
  & \textbf{Project goal: egocentric life assistant with wearable AI glasses} \\
  & The abstract must state the overarching objective of EgoLife: building an AI-powered egocentric life assistant that accompanies users and improves personal efficiency through wearable glasses. \\
\midrule
Introduction
  & \textbf{Vision and motivation for life-oriented egocentric AI assistance} \\
  & The introduction should open with a motivating vision of an AI assistant embedded in daily life that provides personalized, long-term assistance. \\
\midrule
Related Work
  & \textbf{Positioning within egocentric dataset evolution} \\
  & The section should situate EgoLife in the broader history of egocentric vision datasets, starting from early foundational collections. \\
\midrule
\shortstack[l]{Benchmark\\Construction}
  & \textbf{Seven-day multimodal data collection in EgoHouse with six participants} \\
  & The section must state that the benchmark is built from a week-long recording of six volunteers living in a custom environment with multimodal sensing. \\
\midrule
Method
  & \textbf{Overall EgoButler architecture with two subsystems} \\
  & The method section must explain EgoButler with its two subsystems: EgoGPT for continuous clip captioning and EgoRAG for retrieval-augmented QA. \\
\midrule
Experiment
  & \textbf{Main benchmark comparison outcome} \\
  & The section should present the unified comparison protocol via EgoButler and the main benchmark comparison outcome. \\
\bottomrule
\end{tabular}
\end{table*}

\subsection{Example of Rubrics}
\Cref{tab:rubric_example} presents a subset of example rubrics for each section.
The base paper is EgoLife~\citep{yang2025egolife}.

\section{Detailed Prompts}
\subsection{Examples of research\_overview.md}
\label{sec_append:overview}
\begin{tcolorbox}[title=research\_overview\_default.md (EgoLife), breakable]
\begin{scriptsize}
\begin{verbatim}
# EgoLife: Research Overview

## Title
**EgoLife: Towards Egocentric Life Assistant**

---

## 1. Motivation
Existing egocentric datasets and benchmarks capture short, single-person activities and
miss week-long, multi-person social dynamics needed for real-life assistance. The field
needs both a long-duration, multimodal, interpersonal dataset and a method capable of
ultra-long-context reasoning.

---

## 2. Key Insight
A life assistant must fuse egocentric visual-audio cues with long-horizon memory to answer
practical, personalized questions.
**Key idea:** Combine omni-modal clip understanding with hierarchical retrieval over
week-long egocentric recordings.

---

## 3. Benchmark Design: EgoLifeQA
EgoLifeQA is a multiple-choice QA benchmark built on the 266-hour EgoLife dataset (6
participants, 7 days), emphasizing ultra-long-term reasoning. It spans five task types:
EntityLog, EventRecall, HabitInsight, RelationMap, and TaskMaster (3,000 QAs total). Each
QA includes a "certificate length" (look-back requirement), with 2,003 questions requiring
>2 hours of context. Evaluation is accuracy over categories, with evidence retrieval from
week-long recordings.

---

## 4. Data Construction Pipeline
- Data sources: Meta Aria glasses (video, audio, IMU, gaze), 15 exo GoPro cameras, and 2
  mmWave devices in an instrumented "EgoHouse."
- Collection: 6 participants cohabiting for 7 days (~8 h/day), synchronized via EgoSync;
  recordings segmented and aligned; primary language Chinese with English translations.
- Filtering/cleaning: Privacy protection via EgoBlur; synchronization and denoising;
  temporal consistency checks.
- Annotation: Transcripts via Whisper + diarization + human review; 5-min narrated clips
  (0.8x speed) -> merged by GPT-4o into dense visual-audio captions, then verified.
- QA creation/QC: ~100K auto-generated QAs per participant -> manually filtered to 500
  each (3K total), with distractors, audio-needed flags, and certificate lengths;
  SRT-based alignment and multi-pass human verification.

---

## 5. Proposed Method: EgoButler
EgoButler integrates EgoGPT (clip-level omni-modal understanding) and EgoRAG (hierarchical
retrieval for long-context QA). EgoGPT builds on LLaVA-OneVision with an audio branch
(Whisper v3) and is instruction-tuned on EgoIT-99K (99K egocentric QAs), plus Day-1
personalization for identity-awareness. EgoRAG constructs multi-level memory (clip/hour/day
summaries) and retrieves top-k evidence for answer generation.

---

## 6. Key Findings
- EgoGPT achieves state-of-the-art on egocentric benchmarks: 75.4 (EgoSchema), 61.4
  (EgoThink), and 33.4 (EgoPlan).
- On EgoLifeQA, personalization boosts average from 33.1 to 36.0 and RelationMap from
  29.6 to 33.6.
- EgoRAG markedly improves long-context QA: >24h certificate length from 25.0 to 35.4;
  6-24h from 26.8 to 38.9.
- Caption quality is critical: human visual-audio captions reach 45.5 avg; audio-only
  models lag (27-28), visual-only better (31-34), best with audio+visual (36.0).
- Benchmark shows 2,003/3,000 QAs need >2h context, validating the need for retrieval
  over week-long memory.

---

## 7. Contributions
- EgoLife: a 266-hour, week-long, multimodal, ego-exo egocentric dataset with dense
  transcripts and visual-audio captions.
- EgoLifeQA: 3,000 long-context, multiple-choice QAs across five life-assistant task
  categories with certificate-length metadata.
- EgoButler: a two-stage system (EgoGPT + EgoRAG) for personalized, long-horizon
  egocentric QA; release of EgoIT-99K for instruction tuning.
- Comprehensive analysis identifying personalization, caption quality, and hierarchical
  retrieval as key drivers of performance.

---

## 8. Takeaway
EgoLife and EgoButler establish a practical path toward egocentric life assistants by
pairing week-long multimodal data with a retrieval-augmented, omni-modal method for
ultra-long-context reasoning.
\end{verbatim}
\end{scriptsize}
\end{tcolorbox}

\begin{tcolorbox}[title=research\_overview\_long.md (EgoLife), breakable]
\begin{scriptsize}
\begin{verbatim}
# EgoLife: Research Overview

## Title
**EgoLife: Towards Egocentric Life Assistant**

---

## 1. Motivation
Egocentric AI assistants promise to augment daily life by recalling past events, tracking
habits, and making personalized recommendations. However, existing egocentric datasets and
benchmarks (e.g., EPIC-KITCHENS, Ego4D) primarily cover short to medium spans, single-person
views, and narrowly scoped activities. They lack week-long coverage, rich interpersonal
dynamics, and consistent multimodal capture -- all crucial for building assistants that can
reason over ultra-long temporal horizons and understand social context.

This work addresses two gaps simultaneously: (1) the lack of a longitudinal, multiperson,
multimodal egocentric dataset, and (2) the absence of a benchmark that evaluates
long-context, life-oriented assistance. It also contributes a method that integrates
clip-level multimodal understanding with scalable long-context retrieval to answer
week-scale questions. Taken together, the dataset, benchmark, and method move toward
practical egocentric life assistants that operate over days, handle audio-visual inputs,
and keep personalized, identity-aware memory.

---

## 2. Key Insight
The core insight is that long-horizon egocentric assistance requires tightly coupled
components: a personalized, omni-modal clip-understanding model and a hierarchical,
time-aware memory system that supports efficient retrieval over week-level video.

- Key idea: Fuse continuous ego-clip captioning (visual+audio) with hierarchical memory
  summarization and retrieval to answer long-context, life-oriented questions that require
  identity awareness and habit/event grounding across days.

---

## 3. Benchmark Design: EgoLifeQA

### 3.1 Overview
EgoLifeQA is a long-context, life-oriented QA benchmark built on EgoLife, covering a week
of shared living among six participants. It evaluates five capability axes -- object/entity
logging, event recall, habit analysis, social relationship understanding, and task-oriented
assistance -- via multiple-choice questions that require retrieving evidence across hours
to days. The benchmark explicitly annotates whether audio is required and the look-back
time ("certificate length").

Goals:
- Evaluate long-term memory and retrieval over week-scale egocentric content.
- Test personalization (identity recognition, social interaction patterns).
- Assess multi-modal integration (audio + video).

### 3.2 Data Collection & Curation
- Data sources: EgoLife dataset (Meta Aria smart glasses egocentric capture; 15 exocentric
  GoPro cameras; 2 mmWave devices). Primary language: Chinese; annotations translated to
  English.
- Initial QA generation: For each participant, "visual-audio captions" (dense narrations
  post-processed by GPT-4o) were fed into GPT-4o with tailored prompts per question type
  to produce ~100K timestamped QA candidates.
- Human-in-the-loop curation: Annotators reviewed candidates synchronized with video (SRT),
  retained only questions requiring >= 5 minutes of look-back, and prioritized longer
  dependencies and high real-world relevance.
- Finalization: 1,000 questions per participant were pruned and refined to 500 per
  participant (3,000 total), with distractors authored for MCQ. Annotators labeled audio
  requirement and certificate length. Quality control included multi-round reviews and
  timestamp verification.
- Annotation protocol:
  - Five categories (EntityLog, EventRecall, HabitInsight, RelationMap, TaskMaster) with
    structured prompts.
  - Multi-choice answers with evidence timestamps.
  - Minimum evidence window: >= 5 minutes before the question timestamp.

### 3.3 Task Definition & Evaluation Protocol
- Input: Question, candidate answers (MCQ), and access to video-derived memory (captions,
  transcripts) plus retrieval.
- Output: Single-choice answer.
- Metrics: Accuracy (per category and overall), breakdown by certificate length (<2h,
  2-6h, 6-24h, >24h) and by audio requirement.
- Splits: 500 QAs per participant (6 participants -> 3,000 total). Evaluation reports
  per-participant and averaged.

---

## 4. Dataset Statistics
EgoLife (data foundation for EgoLifeQA):
- Participants: 6 volunteers cohabiting for 7 days in a fully instrumented house
  (EgoHouse).
- Recording devices:
  - Egocentric: Meta Aria glasses (video, audio, IMU, gaze; synchronized).
  - Exocentric: 15 GoPro cameras for multi-view coverage.
  - Additional sensing: 2 mmWave devices; house 3D reconstruction with Aria Multi-MPS.
- Duration:
  - ~300 hours captured; 266 hours retained post-cleaning.
  - Approximately 8 hours/day/person; minimum 6h/day enforced.
- Language: Primarily Chinese; English translations provided.
- Annotations:
  - Transcripts: Diarized, speaker-resolved; cross-participant overlap curated (initial
    merged ~50h reviewed, then split and refined).
  - Dense narrations: 361K phrases (avg 2.65 s), captured at 0.8x speed for high density.
  - Merged captions: 25K coherent segments (via GPT-4o-mini merging).
  - Visual-audio captions: Enriched with 1 FPS frames + transcripts, summarized and
    human-verified.
- EgoLifeQA:
  - 3,000 QAs total (500 per participant).
  - Certificate length distribution includes 997 QAs with look-back <2h and 2,003 QAs
    >2h (many beyond 24h).
  - Audio requirement is labeled per QA.

Comparisons:
- Compared to EgoSchema, HourVideo, etc., EgoLife/EgoLifeQA uniquely target week-long,
  multiperson daily life with rich ego-exo, multimodal capture and social dynamics.

---

## 5. Proposed Method: EgoButler

### 5.1 Overview
EgoButler integrates:
- EgoGPT (System-I): A 7B vision-audio-language model for clip-level understanding and
  captioning, adapted to egocentric domain and personalized identity cues.
- EgoRAG (System-II): A hierarchical, time-aware retrieval-augmented generation system
  that builds a multi-level memory bank (clip/hour/day) and retrieves evidence for
  ultra-long-context QA.

Pipeline:
1) Continuous 30 s clip captioning (visual + audio) -> memory bank.
2) Hierarchical retrieval over day/hour windows -> top-k clip evidences.
3) Answer generation with retrieved context.

### 5.2 EgoGPT (Clip-Level Omni-Modal Understanding)
- Base: LLaVA-OneVision (Qwen2-based, 7B).
- Audio branch: Added following Ola-style design; Whisper Large v3 encodes audio; trained
  audio projection on LibriSpeech; finetuned jointly.
- Training data: EgoIT-99K (Table: 9 egocentric datasets; 43.16 h; 1,529 videos incl.
  686 with audio; 99.48K QAs across VC/AVC/MCQ/MRC/IQA).
- Personalization: Additional finetuning on EgoLife Day-1 to calibrate identity cues and
  environment context.
- Functions:
  - Dense visual-audio captioning for 30 s clips (1 FPS sampling in experiments for
    captioners).
  - Answering questions with retrieved evidence and identity-aware references.

### 5.3 EgoRAG (Long-Context Retrieval-Augmented Generation)
- Memory bank M = {(c_i, d_i, t_i)} where:
  - c_i: clip features; d_i: textual descriptions (EgoGPT captions); t_i: hierarchical
    summaries (hourly/day).
- Retrieval:
  - Coarse-to-fine: first retrieve by day -> hour summaries, then fine-grained clip
    selection.
  - Relevance score: s_i = Similarity(q, c_i) + lambda * Similarity(q, d_i). In main
    experiments lambda = 0 (text-only retrieval).
  - Select top-k clips (k typically small; e.g., 3) for evidence.
- Response:
  - r = LLM(q, R), where LLM is EgoGPT or a stronger generator (e.g., GPT-4o)
    conditioned on retrieved evidence R.

---

## 6. Experimental Results

### 6.1 Setup
- Baselines (egocentric VLM benchmarks): GPT-4v/4o, Gemini-1.5-Pro, LLaVA-Next-Video,
  LongVA, IXC-2.5, InternVideo2, Qwen2-VL, Oryx, LLaVA-OV, LLaVA-Videos.
- Datasets: EgoSchema, EgoPlan-Bench, EgoThink; EgoLifeQA (Jake's 500-QA split in this
  version).
- Inputs: 32 frames for standard benchmarks; 1 FPS for caption-based memory in EgoLifeQA.
- EgoRAG: text-similarity retrieval (lambda=0), top-3 30 s clips; re-query with
  GPT-4o-mini; final answer generation by GPT-4o for fairness across captioners.

### 6.2 Key Findings
- EgoGPT on benchmarks (32 frames):
  - EgoGPT (EgoIT-99K): EgoSchema 73.2, EgoPlan 32.4, EgoThink 61.7.
  - With personalization (+D1): EgoSchema 75.4 (+2.2), EgoPlan 33.4 (+1.0), EgoThink
    61.4.
- EgoLifeQA (average accuracy):
  - Gemini-1.5-Pro: 36.9; GPT-4o: 36.2; LLaVA-OV: 30.8.
  - EgoGPT (non-personalized): 33.1; EgoGPT (+D1): 36.0 (+2.9).
  - Category-wise (EgoGPT +D1): EntityLog 39.2, EventRecall 36.5, HabitInsight 31.1,
    RelationMap 33.6, TaskMaster 39.7.
- EgoRAG effectiveness by certificate length:
  - EgoGPT vs. EgoGPT+EgoRAG:
    - <2h: 28.2 -> 27.2;
    - 2-6h: 29.1 -> 35.7;
    - 6-24h: 26.8 -> 38.9;
    - >24h: 25.0 -> 35.4.
  - Large gains for >= 2h look-back validate long-context retrieval benefits.
- Caption quality matters (avg accuracy):
  - Narration-only: 31.5; Transcript-only: 29.6; Visual-audio captions (human-verified):
    45.5.
  - EgoGPT memory banks: audio-only 27.2-28.1; visual-only 31.2-33.6; visual+audio
    33.1-36.0.

---

## 7. Analysis & Insights
- Personalization helps: Day-1 finetuning improves entity and relation questions,
  indicating identity cues and environment priors are beneficial -- but can lead to
  overfitting (e.g., misidentifying later people wearing similar colors).
- Audio is useful but insufficient alone: audio-only captioning underperforms; best
  results require joint visual+audio.
- Long-context retrieval is crucial: naive segmentation causes hallucinations; hierarchical
  memory and retrieval yield pronounced gains especially for >6h queries.
- Error modes:
  - Speech/emotion understanding is limited by ASR-centric audio training (e.g.,
    laughter/emotion recognition).
  - Single-pass retrieval lacks iterative reasoning; failures cascade when relevant clips
    aren't surfaced.
  - Social interaction nuance (e.g., subtle nonverbal cues) remains challenging.

---

## 8. Contributions
- EgoLife dataset: 266 h retained from ~300 h of week-long, multiperson egocentric
  recordings with synchronized ego-exo capture (15 exo cameras, 2 mmWave), rich
  transcripts and dense visual-audio captions.
- EgoLifeQA benchmark: 3,000 long-context MCQs across five categories, with certificate
  lengths up to >24h and audio-requirement labels, targeting practical life assistance.
- EgoButler system:
  - EgoGPT: a 7B vision-audio-language model tuned on EgoIT-99K and personalized on
    EgoLife Day-1, achieving strong egocentric benchmark performance.
  - EgoRAG: hierarchical, time-aware memory and retrieval for ultra-long-context QA,
    significantly improving accuracy for long look-back queries.

---

## 9. Limitations & Future Work
- Dataset scope: single week, six participants, primarily Chinese language; future
  expansions to more participants, diverse locales, and languages are planned.
- Audio understanding: current training biases toward ASR; future models should integrate
  paralinguistic cues (emotion, prosody).
- Personalization stability: day-1 finetuning can overfit identity cues; future work on
  robust identity modeling and lifelong adaptation is needed.
- Retrieval reasoning: EgoRAG is single-pass without multi-hop refinement; integrating
  step-by-step, iterative retrieval and self-correction could improve robustness.
- Privacy and deployment: despite anonymization tools (EgoBlur), real-world deployment
  requires stronger privacy-preserving methods and secure on-device processing.

---

## 10. Takeaway
EgoLife establishes the first week-scale, multiperson, multimodal egocentric dataset and a
long-context QA benchmark targeting life assistance. Coupled with EgoButler -- integrating
personalized clip-level multimodal understanding (EgoGPT) and hierarchical long-context
retrieval (EgoRAG) -- the project advances the frontier of egocentric AI toward practical,
personalized assistants that reason across days, track habits, and understand social
dynamics.
\end{verbatim}
\end{scriptsize}
\end{tcolorbox}

\subsection{Prompts for Creating Overviews}
\begin{tcolorbox}[title=Prompt for Generating Research Overview (Default / Method Paper), breakable]
\begin{scriptsize}
\begin{verbatim}
[System] You are an expert AI researcher. Your task is to read a research paper and
generate a structured research overview in Markdown format. The overview should be
comprehensive yet concise, capturing the key aspects of the paper. It will be used as
input for an AI system that writes LaTeX papers, so accuracy and clarity are critical.
Output ONLY the Markdown content, no preamble or explanation.

[User] Read the following research paper and generate a concise research overview in
Markdown format. Keep it SHORT and focused. Each section should be 2-4 sentences at
most. Total length should be around 1500-2500 characters.

Follow this structure:
# [Paper Acronym/Name]: Research Overview
## Title
## 1. Motivation         (1-2 sentences)
## 2. Key Insight        (1-2 sentences, bold "Key idea:" one-liner)
## 3. Proposed Method     (2-4 sentences)
## 4. Experimental Results (3-5 bullet points with numbers)
## 5. Contributions       (3-4 bullet points)
## 6. Takeaway            (one sentence)
\end{verbatim}
\end{scriptsize}
\end{tcolorbox}

\begin{tcolorbox}[title=Prompt for Generating Research Overview (Long / Method Paper), breakable]
\begin{scriptsize}
\begin{verbatim}
[System] (same as above)

[User] Read the following research paper and generate a detailed and comprehensive
research overview in Markdown format. Be thorough: include technical details, formulas,
specific numbers, and nuanced analysis. Total length should be around 4000-8000
characters.

Follow this structure:
# [Paper Acronym/Name]: Research Overview
## Title
## 1. Motivation          (2-3 paragraphs)
## 2. Key Insight          (bold "Key idea:" summary)
## 3. Proposed Method
    ### 3.1 Overview
    ### 3.2 [Component Name]  (add subsections as needed)
## 4. [Additional Section as needed]
## 5. Experimental Results
    ### 5.1 Setup
    ### 5.2 Key Findings
## 6. Analysis & Insights
## 7. Contributions
## 8. Limitations & Future Work
## 9. Takeaway
\end{verbatim}
\end{scriptsize}
\end{tcolorbox}

\begin{tcolorbox}[title=Prompt for Generating Research Overview (Default / Benchmark Paper), breakable]
\begin{scriptsize}
\begin{verbatim}
[System] (same as above)

[User] Read the following research paper about a benchmark/dataset and generate a
concise research overview in Markdown format. Keep it SHORT and focused. Each section
should be 2-4 sentences at most. Total length should be around 1500-2500 characters.

Follow this structure:
# [Benchmark Name]: Research Overview
## Title
## 1. Motivation
## 2. Key Insight
## 3. Benchmark Design     (tasks, data sources, scale, evaluation protocol)
## 4. Data Construction Pipeline
    (data sources; collection process; filtering/cleaning; annotation protocol;
     number of annotators and qualifications; inter-annotator agreement;
     quality control measures; automated/human-in-the-loop curation steps)
## 5. Key Findings          (3-5 bullet points with numbers)
## 6. Contributions
## 7. Takeaway
\end{verbatim}
\end{scriptsize}
\end{tcolorbox}

\begin{tcolorbox}[title=Prompt for Generating Research Overview (Default / Both Paper), breakable]
\begin{scriptsize}
\begin{verbatim}
[System] (same as above)

[User] Read the following research paper that contributes BOTH a new method AND a new
benchmark/dataset. Generate a concise research overview in Markdown format. Keep it
SHORT and focused. Each section should be 2-4 sentences at most. Total length should be
around 1500-2500 characters.

Follow this structure:
# [Paper Acronym/Name]: Research Overview
## Title
## 1. Motivation
## 2. Key Insight
## 3. Benchmark Design     (tasks, data sources, scale, evaluation protocol)
## 4. Data Construction Pipeline
    (data sources; collection process; filtering/cleaning; annotation protocol;
     quality control measures)
## 5. Proposed Method       (2-4 sentences, key components)
## 6. Key Findings          (3-5 bullet points covering both method and benchmark)
## 7. Contributions
## 8. Takeaway
\end{verbatim}
\end{scriptsize}
\end{tcolorbox}

\begin{tcolorbox}[title=Prompt for Generating Research Overview (Long / Both Paper), breakable]
\begin{scriptsize}
\begin{verbatim}
[System] (same as above)

[User] Read the following research paper that contributes BOTH a new method AND a new
benchmark/dataset. Generate a detailed and comprehensive research overview in Markdown
format. Be thorough: include technical details, formulas, dataset statistics, specific
numbers, and nuanced analysis. Total length should be around 4000-8000 characters.

Follow this structure:
# [Paper Acronym/Name]: Research Overview
## Title
## 1. Motivation             (2-3 paragraphs)
## 2. Key Insight
## 3. Benchmark Design
    ### 3.1 Overview
    ### 3.2 Data Collection & Curation
        (data sources; collection process; filtering/cleaning; annotation protocol;
         annotator count and qualifications; inter-annotator agreement;
         quality control; automated/human-in-the-loop curation)
    ### 3.3 Task Definition & Evaluation Protocol
## 4. Dataset Statistics
## 5. Proposed Method
    ### 5.1 Overview
    ### 5.2 [Component Name]  (add subsections as needed)
## 6. Experimental Results
    ### 6.1 Setup
    ### 6.2 Key Findings
## 7. Analysis & Insights
## 8. Contributions
## 9. Limitations & Future Work
## 10. Takeaway
\end{verbatim}
\end{scriptsize}
\end{tcolorbox}

\begin{tcolorbox}[title=Prompt for Generating Research Overview (Long / Benchmark Paper), breakable]
\begin{scriptsize}
\begin{verbatim}
[System] (same as above)

[User] Read the following research paper about a benchmark/dataset and generate a
detailed and comprehensive research overview in Markdown format. Be thorough: include
technical details, dataset statistics, specific numbers, and nuanced analysis. Total
length should be around 4000-8000 characters.

Follow this structure:
# [Benchmark Name]: Research Overview
## Title
## 1. Motivation            (2-3 paragraphs)
## 2. Key Insight
## 3. Benchmark Design
    ### 3.1 Overview
    ### 3.2 Data Collection & Curation
        (data sources and selection criteria; collection process; filtering/cleaning;
         annotation protocol; annotator count and qualifications; inter-annotator
         agreement; quality control; automated/human-in-the-loop curation)
    ### 3.3 Task Definition & Evaluation Protocol
    ### 3.4 [Additional Design Aspect]
## 4. Dataset Statistics
## 5. Baseline Evaluation
    ### 5.1 Evaluated Methods
    ### 5.2 Key Findings
## 6. Analysis & Insights
## 7. Contributions
## 8. Limitations & Future Work
## 9. Takeaway
\end{verbatim}
\end{scriptsize}
\end{tcolorbox}

\subsection{Prompts for Creating Rubrics}
\begin{tcolorbox}[title=Prompt for Extracting Evaluation Points (Rubric) from Each Section, breakable]
\begin{scriptsize}
\begin{verbatim}
You are an expert paper reviewer. Given a section of an accepted top-tier conference
paper, extract the key elements that are essential for reproducing this section's content.

Each element should represent a distinct piece of information, claim, technical detail,
or structural component that a generated version of this section MUST include to be
considered faithful.

Guidelines:
- Extract 5-15 elements per section depending on length and complexity.
- For Introduction: problem motivation, research gap, proposed approach overview, key
  contributions, paper structure.
- For Method: each major component/module, key equations/formulations, design choices
  and justifications, training procedures.
- For Experiment: datasets, baselines, evaluation metrics, main results, ablation
  studies, key findings.
- For Related Work: main research areas covered, key distinctions from prior work.
- For Abstract: core problem, approach, key results.
- importance should be "high" for elements without which the section would be
  fundamentally incomplete, "medium" for important but not critical details, "low" for
  nice-to-have elements.
- evidence should be a brief quote or reference to the specific part of the GT text.

Respond in JSON format matching the schema exactly. Write element and description in
English. Write evidence in the original language of the paper (usually English).
\end{verbatim}
\end{scriptsize}
\end{tcolorbox}

\subsection{Writing Prompts}
For the writing prompt, we assumed the writing process would be used as part of a practical AI Scientist system, and therefore referred the prompt from Jr. AI Scientist~\citep{miyai2025jr}.

\begin{tcolorbox}[title=Writeup Agent Prompt, breakable]
\begin{scriptsize}
\begin{verbatim}
Your goal is to write up the following idea:

```markdown
{research_overview_text}
```
Note that idea_text represents a preliminary hypothesis and may not necessarily align
with the experiments that were eventually performed.

First, make sure to refer to the experiment data contained in table/ and figure/ folders.

We have VLM-based table descriptions:
```
{table_descriptions}
```

We also have VLM-based figure descriptions:
```
{plot_descriptions}
```

To better understand the methodology and experiments, please also refer to:
- code/ as the proposed method's code implementation

Please read the current template.tex file and update it to produce a complete, coherent,
and scientifically accurate paper.
This must be an acceptable complete LaTeX writeup, suitable for a {num_page}-page
{column_type} paper.
Make sure to use the citations from the references.bib file and report results accurately
based on the experimental data provided.
IMPORTANT: references.bib can be very large. Do NOT read the entire file at once. Instead,
use Grep to search for relevant citation keys or authors, then Read only the specific
portions you need (using offset and limit parameters).
Start by reading template.tex to understand the current state, then edit it to incorporate
all the information above into a complete paper.

Please note: For the bibliography, do not use the \begin{filecontents}{references.bib}
environment. Instead, all citations should refer to an external file named references.bib.
\end{verbatim}
\end{scriptsize}
\end{tcolorbox}

\begin{tcolorbox}[title=Reflection Prompt (LaTeX Error Fixing), breakable]
\begin{scriptsize}
\begin{verbatim}
Now let's reflect and identify any issues (including but not limited to).
Your task is to read the current template.tex file and improve it based on the feedback
provided below.
1) Are there any LaTeX syntax errors or style violations we can fix? Refer to the chktex
output below.

chktex results:
```
{check_output}
```

2) Are there any LaTeX compilation errors? Refer to the tectonic compile output below.

tectonic compile output:
```
{compile_output}
```

If there are errors reported above, please fix them directly.
Read template.tex and edit it to address these issues.
Focus especially on fixing compilation errors so that the paper compiles successfully.

If no errors are reported, no changes are necessary.
\end{verbatim}
\end{scriptsize}
\end{tcolorbox}

\begin{tcolorbox}[title=Page Limit Adjustment Prompt, breakable]
\begin{scriptsize}
\begin{verbatim}
The main text (before 'References') is currently {main_pages} pages. The target is
{page_limit} pages.

The paper is {status}. Please {action} to reach the target.
Do NOT move content to or create an Appendix. Keep everything in the main text.
Do not add or remove more than 1000 characters in this revision. Do not use
\begin{filecontents}{references.bib}.
\end{verbatim}
\end{scriptsize}
\end{tcolorbox}

\subsection{Evaluation Prompts}
\begin{tcolorbox}[title=Rubric Evaluation: Section-Level Scoring, breakable]
\begin{scriptsize}
\begin{verbatim}
[System Message]
You are an expert paper reviewer. You are given:
1. A list of key elements (rubric) that should appear in a specific section of
   a paper.
2. The predicted section content to evaluate.
3. (Optional) Figure/Table context showing which visual assets are present or
   missing.

For each element, score how well the predicted section covers it on a 1-5 scale:
5: Fully and accurately described. The element is present with correct details.
4: Mostly described. The core idea is present but some details are missing or
   slightly imprecise.
3: Partially described. The element is mentioned but with significant gaps or
   vagueness.
2: Barely mentioned. Only a superficial or indirect reference exists.
1: Not described at all. The element is completely absent from the predicted
   section.

Respond in JSON format. For each element, provide:
- "element": the element name (copy exactly from input)
- "score": 1-5
- "reasoning": brief explanation

[User Message]
**Section: {section_name}**

**Rubric (key elements to check):**
- {element_name} ({importance}): {description}
- ...

**Predicted section content:**
{pred_content}

### Figure/Table Context for this Section
The following analysis describes the presence or absence of visual assets
in this section compared to the Ground Truth:
{figure_table_context}

**Instructions:**
1. Evaluate each rubric element based on the text content.
2. CRITICAL: If a rubric element requires or refers to data/visuals, and the
   corresponding Figure/Table is reported as MISSING or has low match score in
   the context above, you MUST penalize the score for that element.
Evaluate each element on a 1-5 scale.
\end{verbatim}
\end{scriptsize}
\end{tcolorbox}

\begin{tcolorbox}[title=Hallucination Evaluation: Stage 1 --- Claim Extraction \& Classification (LLM), breakable]
\begin{scriptsize}
\begin{verbatim}
[System Message]
You are an expert paper reviewer detecting factual errors in a predicted paper
section.

You are given:
1. The predicted section content.
2. The ground truth (GT) full paper content for reference.

Your task is to identify all concrete, verifiable claims in the predicted section
(e.g., specific numbers, method descriptions, experimental setups, results)
and classify each into one of three categories:

- supported: The claim is directly stated in or logically derivable from the GT
  paper.
- neutral: The claim is NOT in the GT, but is a reasonable general statement,
  common knowledge, or supplementary detail that does not contradict the GT.
  This is NOT an error.
- contradictory: The claim directly contradicts specific information in the GT
  paper. This is a factual error / hallucination.

For contradictory claims, also assign severity:
- major: Incorrect numbers, fabricated results, wrong method descriptions,
  misattributed findings -- errors that would mislead a reader.
- minor: Overly strong generalizations, imprecise wording that slightly distorts
  meaning, minor numerical rounding issues.

IMPORTANT:
- Do NOT classify claims as contradictory simply because they are absent from the
  GT. Absence != contradiction.
- Focus on claims that can be verified against the GT. Skip purely stylistic or
  structural observations.
- Be thorough: extract ALL verifiable claims, not just a few.

Respond in JSON with a list of claims. Each claim has:
- "claim": the specific statement from Pred
- "classification": "supported" | "neutral" | "contradictory"
- "evidence": brief explanation
- "severity": "major" | "minor" (only for contradictory, null otherwise)

[User Message]
**Section being evaluated: {section_name}**

**Predicted section content:**
{pred_content}

**Ground Truth full paper (for reference):**
{gt_full_content}

Extract and classify all verifiable claims from the predicted section.
\end{verbatim}
\end{scriptsize}
\end{tcolorbox}

\begin{tcolorbox}[title=Hallucination Evaluation: Stage 2 --- Agent-Based Verification, breakable]
\begin{scriptsize}
\begin{verbatim}
[System Message]
You are a rigorous fact-checker performing a second-pass verification.

A previous reviewer flagged the following claim as contradictory (factual error /
hallucination) in a predicted paper section. Your job is to carefully re-examine
whether this is truly a contradiction with the Ground Truth paper, or a false
positive.

Classify the claim into one of:
- contradictory: Confirmed. The claim genuinely contradicts specific information
  in the GT paper.
- neutral: False positive. The claim is absent from the GT but does NOT
  contradict it. Absence is not contradiction.
- supported: False positive. The claim is actually supported by the GT paper.

For confirmed contradictory claims, assign severity:
- major: Incorrect numbers, fabricated results, wrong method descriptions.
- minor: Overly strong generalizations, imprecise wording.

Respond in JSON:
- "classification": "supported" | "neutral" | "contradictory"
- "severity": "major" | "minor" for contradictory, "none" for supported/neutral
- "evidence": brief explanation of why you changed or kept the classification

[User Message]
The following {N} claims were flagged as contradictory by a previous reviewer.
Re-examine EACH claim and classify it independently.

### Claim 1
- Claim: {claim_text}
- Original evidence: {original_evidence}
- Original severity: {original_severity}
...

Your current working directory contains the Ground Truth paper resources:
- The GT paper's LaTeX source (main.tex, gt_main.tex, or similar .tex files)
- code/ directory with the original codebase (if exists)
- figures/ directory with figure images
- tables/ directory with table data (.tex files)

Please read the relevant GT files to verify each claim. For claims about the
implementation details or methods, check the code/ directory. For other claims,
you can check main.tex/gt_main.tex and tables/.
Return a JSON with a "results" array containing one entry per claim, in the same
order.

Note: The agent operates in READ-ONLY mode with access to Read, Glob, and Grep
tools only.
\end{verbatim}
\end{scriptsize}
\end{tcolorbox}

\section{Generated Papers}
We include the two generated papers based on PINO~\citep{ota2025pino}.



\section*{Supplementary material}

# PINO: Person-Interaction Noise Optimization for Long-Duration and Customizable Motion Generation of Arbitrary-Sized Groups

Sakuya Ota    Qing Yu    Kent Fujiwara    Satoshi Ikehata    Ikuro Sato

## Abstract

*Generating realistic multi-person interactions at scale remains an open challenge in human motion synthesis. Existing methods either rely on shared text prompts that yield uniform, undifferentiated behaviors, or require retraining when new constraints are introduced. We present PINO (**P**erson-**I**nteraction **N**oise **O**ptimization), a training-free framework that composes arbitrary-sized group interactions from a pretrained two-person diffusion model (InterGen). Our key insight is to decompose group dynamics into a sequence of pairwise generations anchored by a shared "pivot" character. For each new person added to the group, PINO conditions the target character on the pivot via masked diffusion and optimizes the initial noise with differentiable physics-based penalties, including overlap avoidance, root position, movement region, facing orientation, and relative positioning. Per-pair text prompts enable fine-grained role assignment without shared-prompt ambiguity. For long-duration sequences, PINO employs motion inpainting with boundary smoothness penalties to extend interactions while preserving temporal coherence. Experiments on the InterHuman dataset demonstrate that PINO eliminates body overlap (Overlap $\approx$ 0.000 vs. 0.119 for InterGen in the two-person case), dramatically reduces penetration volume (275.65 vs. 3112.72), and maintains competitive semantic quality (FID 13.163 vs. 13.278). These benefits scale to groups of five or more persons without any additional training.*

## 1. Introduction

Synthesizing realistic human motion has broad applications in computer animation, video games, and robotics. While single-person text-to-motion generation has achieved impressive results [1, 3, 20, 26], extending these techniques to multi-person interactions poses unique challenges: motions must be jointly plausible, semantically consistent with natural language descriptions, and physically coherent (i.e., free of body penetration and interpenetration artifacts).

Existing approaches to multi-person generation fall into three broad categories. First, methods based on Control-Net [25] (e.g., InterControl [22], FreeMotion [2]) adapt single-person or two-person networks through conditional guidance. Second, directly trained multi-person networks (e.g., Shan et al. [16]) are learned on multi-person datasets, limiting scalability. Across all categories, a common limitation is that adding new characters or introducing new spatio-temporal constraints typically requires either retraining or significant model redesign.

Inspired by Diffusion Noise Optimization (DNO) [7] and ProgMoGen [11], which demonstrated that optimizing the initial noise of a diffusion model—rather than its weights—enables zero-shot controllability for single-person motion, we extend this principle to the multi-person setting. Our approach, PINO (**P**erson-**I**nteraction **N**oise **O**ptimization), leverages a pretrained two-person diffusion model (InterGen [10]) and composes arbitrarily large groups through sequential, pairwise generation guided by physics-aware noise optimization.

The core idea of PINO is to treat group interactions as a chain of pairwise relationships anchored by a pivot character. When generating person $k$ ($k > 2$), we fix the motion of the pivot (person 1) via feature-space masking and generate person $k$ conditioned on the pivot using a per-pair text prompt. The initial noise for person $k$ is optimized through a differentiable loss that combines: (i) an *overlap penalty* enforcing minimum distance between all character pairs; (ii) a *root position penalty* controlling start/end positions; (iii) a *movement region penalty* constraining spatial extents; (iv) a *facing orientation penalty* specifying relative body orientations; and (v) a *relative position penalty* governing pairwise spatial relationships. For long-duration generation, motion inpainting with a boundary smoothness penalty enables coherent sequence extension without retraining.

Our contributions are as follows:

- A **training-free** framework that scales a pretrained two-person diffusion model to arbitrary group sizes via sequential pairwise generation with pivot conditioning.

- **Physics-aware noise optimization** that enforces overlap avoidance and fine-grained spatio-temporal controls entirely at inference time.

- **Per-pair prompt switching** that enables nuanced role assignment and customizable behaviors within large groups, avoiding shared-prompt ambiguity.
- **Long-duration motion generation** via inpainting with smooth boundary transitions, enabling evolving interactions over time.

Experiments on the InterHuman benchmark show that PINO achieves near-zero overlap even in five-person scenarios, dramatic reductions in penetration volume, and competitive or superior FID scores compared to InterGen and FreeMotion, all without additional training.

## 2. Related Work

**Single-Person Motion Generation.** Text-conditioned human motion synthesis has made rapid progress using variational autoencoders [12, 13], transformer-based architectures [3], and diffusion models [1, 20, 26]. MDM [20] introduced a classifier-free diffusion model for motion that predicts the clean sample at each denoising step. MLD [1] performs diffusion in a learned latent space to reduce computational cost. These works focus on single-person scenarios and serve as the backbone components for multi-person extensions.

**Controllable Motion Generation.** Adding spatial constraints to diffusion models without retraining is an active research direction. GMD [6] introduces imputation-based spatial guidance for pelvis trajectory control. OmniControl [23] extends this to any joint at any time. ProgMoGen [11] demonstrates that open-set motion control tasks can be solved by optimizing the diffusion latent code using programmable constraint functions—inspiring our noise optimization approach. DNO [7] further formalizes noise optimization as a universal prior for single-person motion editing, completion, and denoising. EDGE [21] proposes in-betweening for dance, a form of motion inpainting that is related to our long-duration extension.

**Multi-Person Interaction Generation.** Classical methods synthesize interactions by retargeting or blending captured motion clips [8, 17]. With the rise of learning-based approaches, more flexible interaction generation has become possible. ==ActFormer [24] uses a GAN-based transformer for multi-person action-conditioned generation.== InterGen [10] introduces a two-person diffusion model with weight-sharing denoisers and mutual attention trained on the InterHuman dataset [10]. PriorMDM [15] explores compositional use of single-person priors through a learned communication block (ComMDM) for two-person interaction. InterHandGen [9] addresses two-hand interaction generation using cascaded reverse diffusion. Role-aware interaction generation [19] distinguishes active and passive

> GT misplaces ActFormer under single-person GAN-based motion generation instead of multi-person interaction generation.

roles via text conditioning. InterMask [5] uses collaborative masked modeling in discrete space for high-fidelity two-person generation. DiffCollage [27] composes large content via factor graph representations of overlapping diffusion models.

**Scaling to Arbitrary Groups.** Going beyond two persons, ==InterControl [22] adapts a single-person diffusion model to multi-person scenarios via zero-shot joint-contact guidance.== FreeMotion [2] unifies single and multi-person synthesis through decoupled generation and interaction modules but still relies on ControlNet adapters. Shan et al. [16] train a multi-person transformer diffusion model on large-scale datasets curated from internet videos. Table 1 compares these methods along key design dimensions. Unlike prior work, PINO requires no retraining, supports per-pair prompts, and achieves physical plausibility through noise optimization.

> GT describes InterControl as ControlNet-based joint guidance, not zero-shot joint-contact guidance.

## 3. Preliminaries

**Two-Person Diffusion Model.** We build on InterGen [10], a denoising diffusion probabilistic model [4] for two-person interaction synthesis. Given a text prompt $c$, InterGen learns the distribution over two-person motion sequences $\mathbf{x} = (\mathbf{x}^{(1)}, \mathbf{x}^{(2)}) \in \mathbb{R}^{T \times 2 \times D}$, where $T$ is the number of frames and $D = 262$ is the per-person feature dimension (encoding joint positions, velocities, rotations, and foot contact). The model uses two weight-sharing transformer denoisers with mutual attention to capture inter-person dependencies. At inference, a DDIM sampler [18] denoises the joint motion from Gaussian noise conditioned on a CLIP [14] text embedding.

**Diffusion Noise Optimization.** DNO [7] and ProgMoGen [11] demonstrated that, given a pretrained diffusion model $\epsilon_\theta$, one can optimize the initial noise $\mathbf{z}_T \sim \mathcal{N}(0, \mathbf{I})$ to satisfy a target objective $\mathcal{L}$ without modifying model weights:

$$\mathbf{z}_T^* = \arg\min_{\mathbf{z}_T} \mathcal{L}(\hat{\mathbf{x}}_0(\mathbf{z}_T)), \qquad (1)$$

Table 1. Comparison of recent multi-person interaction generation methods, including single-person method ProgMoGen [11].

| Method | #Person in Network Design | #Person in Training Data | Order of Multi-Person Generation | Method |
|---|---|---|---|---|
| InterControl [22] | One | One | Sequential | ControlNet |
| FreeMotion [2] | One | Two | Sequential | ControlNet |
| Shan et al. [16] | Multiple | Multiple | Concurrent | Supervised Training |
| ProgMoGen [11] | One | Training-free | N/A (Single-Person) | Noise Optimization |
| Ours | Two (or multiple) | Training-free | Sequential | Noise Optimization |

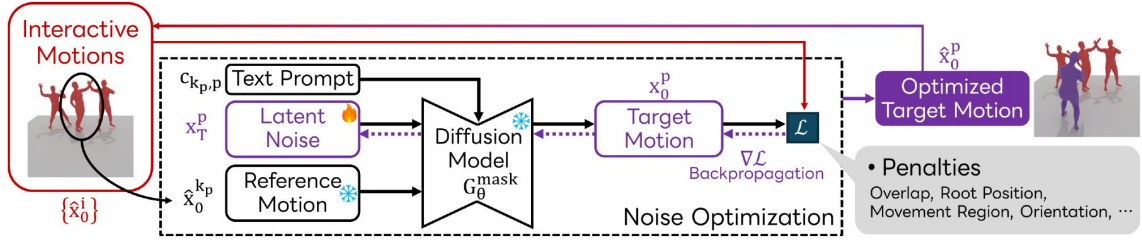

Figure 1. **Person-Interaction Noise Optimization (PINO).** Starting from a two-person generation conditioned on a per-pair prompt, PINO adds characters incrementally. For each new person, person 1 acts as the pivot (reference): its motion is injected into the initial noise via feature masking, and the new person's noise is optimized with physics-based penalties. The same procedure is repeated for long-duration extension via motion inpainting.

where $\hat{\mathbf{x}}_0(\mathbf{z}_T) = \text{Denoise}_\theta(\mathbf{z}_T, c)$ is the sample generated by the full reverse diffusion process from $\mathbf{z}_T$. Gradients flow through the entire denoising chain back to $\mathbf{z}_T$. This enables arbitrary differentiable constraints to be imposed at inference time.

## 4. Method

### 4.1. Overview

PINO generates group interactions of arbitrary size in a two-stage pipeline (Fig. 1):

1. **Base generation:** A two-person interaction (persons 1 and 2) is generated with InterGen conditioned on a text prompt $c_{12}$, with noise optimized to satisfy physics-based penalties.

2. **Sequential extension:** Each additional person $k$ ($k \geq 3$) is generated conditioned on person 1 (the pivot) using a per-pair prompt $c_{1k}$, with its noise optimized to avoid overlap with all previously generated persons.

For long-duration generation, a third stage extends any two-person sequence using motion inpainting with boundary smoothness penalties.

### 4.2. Masked Pivot Conditioning

When generating person $k$ ($k \geq 3$) in interaction with the pivot (person 1), we fix person 1's motion $\mathbf{x}^{(1)} \in \mathbb{R}^{T \times D}$ and generate a new pair $(\tilde{\mathbf{x}}^{(1)}, \mathbf{x}^{(k)})$ using InterGen. The key is to constrain $\tilde{\mathbf{x}}^{(1)}$ to match the existing pivot motion. Following the inpainting strategy in the diffusion literature, we construct the initial noise as:

$$\mathbf{z}_T = \mathbf{m} \odot \mathbf{x}^{(1)} + (1 - \mathbf{m}) \odot \boldsymbol{\epsilon}, \quad \boldsymbol{\epsilon} \sim \mathcal{N}(0, \mathbf{I}), \quad (2)$$

where $\mathbf{m}$ is a binary mask selecting the feature dimensions corresponding to person 1. During each denoising step, the predicted person 1 features are replaced by the fixed pivot motion, ensuring the generated pair remains anchored to the existing group. Person $k$'s noise is left free to be optimized.

### 4.3. Physics-Aware Noise Optimization

Starting from the masked initial noise, we optimize the free (person $k$) portion of $\mathbf{z}_T$ by minimizing a composite physics-based loss:

$$\mathcal{L} = \mathcal{L}_{\text{overlap}} + \lambda_r \mathcal{L}_{\text{root}} + \lambda_{\text{reg}} \mathcal{L}_{\text{region}} + \lambda_o \mathcal{L}_{\text{orient}} + \lambda_{\text{rel}} \mathcal{L}_{\text{rel}}, \quad (3)$$

where $\lambda_r, \lambda_{\text{reg}}, \lambda_o, \lambda_{\text{rel}}$ are scalar weights. We describe each penalty below.

**Overlap Penalty** ($\mathcal{L}_{\text{overlap}}$). Body interpenetration is a common artifact in multi-person generation. We penalize configurations in which any two characters' root (pelvis) positions on the XZ plane are closer than a minimum distance $d_{\min}$ or farther than $d_{\max}$:

$$\mathcal{L}_{\text{overlap}} = \frac{1}{T} \sum_{t=1}^{T} \left[ \max(0, d_{\min} - \|\mathbf{p}_1^t - \mathbf{p}_k^t\|) + \max(0, \|\mathbf{p}_1^t - \mathbf{p}_k^t\| - d_{\max}) \right] \quad (4)$$

where $\mathbf{p}_i^t \in \mathbb{R}^2$ is the XZ position of the pelvis of person $i$ at frame $t$. When $k > 3$, we additionally penalize person $k$ against all previously generated persons (excluding the pivot) using a separate distance threshold $d_{\text{other}}$.

**Root Position Penalty** ($\mathcal{L}_{\text{root}}$). To control start and/or end positions, we define a target root position $\mathbf{g}_i \in \mathbb{R}^2$ for person $i$ and penalize deviation:

$$\mathcal{L}_{\text{root}} = \max(0, \|\mathbf{p}_i^\tau - \mathbf{g}_i\|^2 - \eta), \quad (5)$$

where $\tau$ is the target time frame (start or end) and $\eta$ is a tolerance margin.

**Movement Region Penalty** ($\mathcal{L}_{\text{region}}$). When a spatial region constraint is specified (e.g., "person must stay within a 6m × 6m rectangle centered at the origin"), we penalize

3

frames where the root position violates the boundary:

$$\mathcal{L}_{\text{region}} = \frac{1}{T} \sum_{t=1}^{T} \mathbb{1}[\mathbf{p}_k^t \notin \mathcal{R}] \cdot \text{dist}(\mathbf{p}_k^t, \mathcal{R}), \quad (6)$$

where $\mathcal{R}$ is a circular or rectangular region and $\text{dist}(\cdot, \mathcal{R})$ is the distance from the boundary.

**Facing Orientation Penalty ($\mathcal{L}_{\text{orient}}$).** To control relative facing directions between persons, we define a target cosine similarity $\hat{s}$ between the forward-facing unit vectors $\mathbf{n}_1^t$ and $\mathbf{n}_k^t$ derived from each person's hip joints:

$$\mathcal{L}_{\text{orient}} = \frac{1}{T} \sum_{t=1}^{T} \max(0, \cos_{\min} - \mathbf{n}_1^t \cdot \mathbf{n}_k^t), \quad (7)$$

where $\cos_{\min}$ is a threshold (e.g., $0.9$) above which the constraint is considered satisfied.

**Relative Position Penalty ($\mathcal{L}_{\text{rel}}$).** To specify the desired relative spatial arrangement between persons (e.g., "person $k$ stands to the left of person 1"), we define a target relative displacement vector $\boldsymbol{\delta}_{1k} \in \mathbb{R}^2$ in the XZ plane and penalize deviation from it:

$$\mathcal{L}_{\text{rel}} = \frac{1}{T} \sum_{t=1}^{T} \max\left(0, \|(\mathbf{p}_k^t - \mathbf{p}_1^t) - \boldsymbol{\delta}_{1k}\|^2 - \eta_{\text{rel}}\right), \quad (8)$$

where $\eta_{\text{rel}}$ is a tolerance margin allowing small deviations. This penalty complements the overlap penalty by not only preventing characters from being too close, but actively guiding their arrangement into semantically meaningful configurations.

**Optimization.** We optimize $\mathbf{z}_T$ using Adam with learning rate $1 \times 10^{-2}$ and early stopping when $\mathcal{L} < \epsilon_{\text{stop}} = 10^{-3}$, running up to $N_{\text{iter}} = 50$ iterations. Loss weights are fixed to $\lambda_r = 10$, $\lambda_{\text{reg}} = \lambda_o = \lambda_{\text{rel}} = 1$ across all experiments and group sizes. Gradients flow through the entire denoising chain by retaining the computation graph during the reverse diffusion. Gradient norms are clipped to $1.0$ before each parameter update to prevent instabilities that arise early in the optimization trajectory, when denoising paths are farthest from the data manifold. The optimization is performed on a single GPU and does not modify any model parameters.

### 4.4. Long-Duration Generation via Inpainting

For long-duration scenarios, we use motion inpainting to extend a completed short sequence $\mathbf{x}_{[0, T_0]}$ by an additional $\Delta T$ frames. The extension shares an overlap window of $T_{\text{ov}}$ frames with the existing sequence; the noise for these

frames is initialized with the existing motion rather than pure Gaussian noise (analogous to Eq. (2) applied to temporal dimensions). An additional boundary smoothness penalty penalizes discontinuities in joint velocities at the transition frame:

$$\mathcal{L}_{\text{smooth}} = \|\dot{\mathbf{x}}_{T_0}^{(i)} - \dot{\mathbf{x}}_{T_0+1}^{(i)}\|^2, \quad (9)$$

where $\dot{\mathbf{x}}_t^{(i)}$ denotes the joint velocity of person $i$ at frame $t$. The overlap penalty from Eq. (4) is also applied throughout the extended sequence to maintain physical plausibility. The extended sequence is obtained by concatenating the original and newly generated segments at the overlap boundary.

### 4.5. Per-Pair Prompt Switching

A key advantage of the sequential pairwise generation framework is that each pair $(1, k)$ can be conditioned on a distinct text prompt $c_{1k}$ describing the specific interaction between the pivot and person $k$. For example, in a five-person group, one might specify:

- $c_{12}$: "Two people shake hands."
- $c_{13}$: "Two people high-five."
- $c_{14}$: "Two people bow to each other."
- $c_{15}$: "Two people dance together."

This fine-grained control is impossible with shared-prompt methods (e.g., InterGen applied to a single joint prompt for all persons), where all characters tend to perform the same

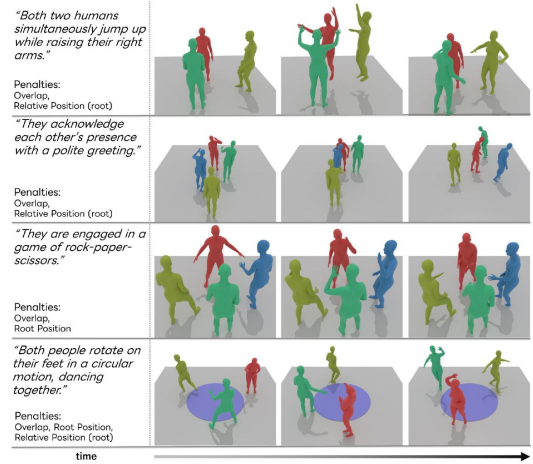

Figure 2. Qualitative results of multi-person interaction generation with PINO. From top to bottom: three-person, four-person, and five-person scenarios. Each group is generated sequentially, with person 1 serving as pivot. Despite increasing group size, characters maintain clear separation and semantically meaningful interactions.

*"Both two humans simultaneously jump up while raising their right arms."*

Penalties:
Overlap,
Relative Position (root)

*"They acknowledge each other's presence with a polite greeting."*

Penalties:
Overlap,
Relative Position (root)

*"They are engaged in a game of rock-paper-scissors."*

Penalties:
Overlap,
Root Position

*"Both people rotate on their feet in a circular motion, dancing together."*

Penalties:
Overlap, Root Position,
Relative Position (root)

time

4

or similar motion. Our ablation study (Section 5.5) confirms the contribution of each loss component, including how the noise optimization enforces different spatial roles.

# 5. Experiments

## 5.1. Experimental Setup

**Dataset.** We evaluate PINO on the InterHuman dataset [10], a large-scale collection of two-person interaction motion sequences (approximately 107M frames across 23,337 text-annotated interactions). For multi-person evaluation we compose two-person pairs from this dataset, and for five-person evaluation we extend the composition chain.

**Baselines.** We compare against:
- **InterGen** [10]: The pretrained two-person diffusion model used as our backbone. For multi-person scenarios, InterGen is applied with inpainting (denoted *InterGen (inpainting)*).
- **FreeMotion** [2]: A ControlNet-based approach that supports number-free multi-person generation.

Our method is denoted **PINO-InterGen** to indicate the use of InterGen as the base model.

**Metrics.** We evaluate along two axes:
- *Non-semantic (physics) metrics*: **Overlap** (fraction of frames with root distance below threshold), **Penetration Volume** (PenVol; cm³, volumetric interpenetration), **Foot Skate** (foot sliding artifact), and **Max Acceleration** (Max Acc.; physical plausibility of dynamics).
- *Semantic metrics*: **FID** (Fréchet Inception Distance, computed on motion features from a pretrained encoder), **Diversity** (Div.), and **R-Precision** (Top-3; motion-text retrieval accuracy).

**Implementation Details.** All experiments use InterGen's pretrained checkpoint without any fine-tuning. DDIM sampling runs for $T_{\text{ddim}} = 50$ steps with a classifier-free guidance scale of 2.5. Noise optimization for each additional person requires a separate gradient-enabled forward pass through the full denoising chain. Distance thresholds for the overlap penalty are set to $d_{\min} = 0.5\,\text{m}$ and $d_{\max} = 2.0\,\text{m}$ for pivot pairs, and $d_{\text{other}} = 0.3\,\text{m}$ for non-pivot pairs. For long-duration generation, the temporal overlap window is fixed at $T_{\text{ov}} = 30$ frames. All hyperparameters are held constant across group sizes and datasets, demonstrating the robustness of PINO to hyperparameter choices.

## 5.2. Two-Person Interaction Generation

Table 2 reports results for two-person interaction generation with overlap avoidance. PINO achieves essentially zero

| Method | Overlap ↓ | PenVol. (cm³) ↓ | Foot Skate ↓ | Max Acc. ↓ |
|---|---|---|---|---|
| GT | 0.029 | 471.75 | 0.099 | 0.074 |
| InterGen | 0.119 | 3112.72 | **0.124** | **0.034** |
| PINO-InterGen | **0.000** | **275.65** | 0.125 | **0.034** |

(a) Non-semantic metrics.

| Method | FID ↓ | Div. → | R-Prec. ↑ (Top3) |
|---|---|---|---|
| GT | 0.983 | 7.921 | 0.715 |
| InterGen | 13.278 | 7.793 | 0.674 |
| PINO-InterGen | **13.163** | **7.904** | **0.675** |

(b) Semantic metrics.

Table 2. Evaluation of two-person interaction generation while avoiding overlap.

| Method | Overlap↓ | PenVol.↓ | Foot Skate↓ | Max Acc.↓ |
|---|---|---|---|---|
| InterGen (inpainting) | 0.766 | 25215.78 | 0.099 | **0.030** |
| PINO-InterGen | **0.000** | **785.01** | **0.091** | 0.031 |

Table 3. Evaluation of multi-person interaction generation

overlap (0.000) and dramatically reduces penetration volume (275.65 cm³) compared to InterGen (0.119 overlap, 3112.72 cm³ PenVol). Remarkably, the semantic quality is preserved or improved: FID drops from 13.278 (InterGen) to 13.163 (PINO-InterGen), and Diversity (7.904 vs. 7.793) and R-Precision (0.675 vs. 0.674) are both slightly better. Ground-truth (GT) statistics are provided as upper bounds on quality. Foot Skate and Max Acceleration remain on par with InterGen, showing that noise optimization does not introduce spurious dynamics.

## 5.3. Multi-Person Interaction Generation

Table 3 compares PINO against InterGen with inpainting for multi-person (three-person and beyond) generation. PINO eliminates overlap entirely (0.000 vs. 0.766) and reduces PenVol by a factor of $\sim 32$ (785.01 vs. 25215.78), while also improving Foot Skate (0.091 vs. 0.099). Max Acceleration is slightly higher for PINO (0.031 vs. 0.030), but the difference is negligible. These results confirm that the sequential pairwise framework with noise optimization effectively prevents physical artifacts as the group grows.

**Scaling to Five Persons.** Table 4 reports pairwise FID and overlap for groups of up to five persons (where pair $(1, k)$ reports metrics for persons 1 and $k$). PINO consistently achieves low overlap across all pairs, reaching only 0.069 for pair $(1, 5)$ compared to 0.977 for InterGen and 0.991 for FreeMotion. FID is also lower for all pairs: e.g., 16.911 vs. 19.501 (InterGen) and 25.671 (FreeMotion) for pair $(1, 5)$. Foot Skate and Max Acceleration are slightly higher than InterGen's in some pairs, a minor trade-off for the dramatic reduction in physical artifacts. Fig. 2 illustrates the diversity of generated group interactions.

5

| Method | Pair | FID↓ | Overlap↓ | Foot Skate↓ | Max Acc.↓ |
|--------|------|------|----------|-------------|-----------|
| FreeMotion | (1,2) | 19.383 | 0.255 | **0.112** | **0.034** |
| | (1,3) | 25.184 | 0.806 | **0.122** | 0.038 |
| | (1,4) | 25.812 | 0.964 | 0.127 | 0.040 |
| | (1,5) | 25.671 | 0.991 | 0.130 | 0.042 |
| InterGen | (1,2) | 13.100 | 0.119 | 0.126 | **0.034** |
| | (1,3) | 19.414 | 0.799 | **0.125** | **0.036** |
| | (1,4) | 19.386 | 0.937 | **0.124** | **0.038** |
| | (1,5) | 19.501 | 0.977 | **0.124** | **0.039** |
| PINO-InterGen | (1,2) | **12.920** | **0.000** | 0.127 | **0.034** |
| | (1,3) | **17.170** | **0.002** | 0.135 | 0.040 |
| | (1,4) | **16.890** | **0.019** | 0.138 | 0.043 |
| | (1,5) | **16.911** | **0.069** | 0.142 | 0.043 |

Table 4. Evaluation of multi-person interaction generation. From pair (1,2), interactions are generated incrementally using 1 as a pivot. Overlap is among all individuals generated up to that step.

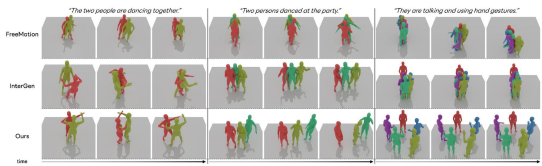

Figure 3. Qualitative comparison of multi-person interaction generation. InterGen (inpainting) produces severe body overlap artifacts. PINO maintains clear separation between all persons across frames.

| Method | Overlap↓ | PenVol.↓ | Foot Skate↓ | Max Acc.↓ |
|--------|----------|----------|-------------|-----------|
| InterGen (inpainting) | 0.672 | 15622.16 | 0.070 | 0.044 |
| PINO-InterGen | **0.031** | **1375.24** | **0.045** | **0.043** |

Table 5. Evaluation of non-semantic elements in motion extension.

## 5.4. Long-Duration Motion Extension

For the motion extension task, we evaluate PINO in two complementary ways.

**Non-semantic evaluation.** Table 5 shows that the extended motions generated by PINO have significantly lower overlap (0.031 vs. 0.672), lower PenVol (1375.24 vs. 15622.16), and lower Foot Skate (0.045 vs. 0.070) and Max Acceleration (0.043 vs. 0.044) compared to InterGen inpainting. The boundary smoothness penalty effectively prevents discontinuities at the transition frame.

**Semantic evaluation.** Table 6 evaluates semantic quality by cropping extended sequences to match ground-truth lengths. PINO achieves a lower FID (18.930 vs. 20.390), zero overlap (0.0 vs. 0.089), lower Foot Skate (0.054 vs. 0.081), and lower Max Acceleration (0.025 vs. 0.036). This confirms that noise-optimized inpainting not only avoids artifacts but also maintains the semantic fidelity of the generated motions.

| Method | FID↓ | Overlap↓ | Foot Skate↓ | Max Acc.↓ |
|--------|------|----------|-------------|-----------|
| InterGen | 20.390 | 0.089 | 0.081 | 0.036 |
| PINO-InterGen | **18.930** | **0.0** | **0.054** | **0.025** |

Table 6. Evaluation of semantic elements in two-person motion extension. Extended motions are cropped to match the ground truth length for evaluation.

| Method | Pos. Err.↓ | Overlap↓ | Reg. Viol.↓ | Orient. Err.↓ |
|--------|-----------|----------|-------------|---------------|
| InterGen | 1.0 | 0.292 | 0.500 | 1.0 |
| + $\mathcal{L}_{\text{root}}$ | **0.0** | 0.333 | 0.917 | 1.0 |
| + $\mathcal{L}_{\text{overlap}}$ | **0.0** | **0.0** | 0.958 | 1.0 |
| + $\mathcal{L}_{\text{region}}$ | 0.083 | **0.0** | **0.043** | 1.0 |
| + $\mathcal{L}_{\text{orientation}}$ | 0.083 | 0.043 | 0.083 | **0.208** |

Table 7. Ablation study on motion penalties. Metrics include positional errors (Pos. Err.↓), overlap (Overlap↓), region violations (Reg. Viol.↓), and orientation errors (Orient. Err.↓). Positional errors are measured with a threshold of 20 cm, region violations with a threshold of 10 cm, and orientation errors with a threshold of 20 degrees. Lower values indicate better performance.

## 5.5. Ablation Study

To understand the contribution of each penalty term, Table 7 reports spatio-temporal control metrics as losses are added incrementally. Starting from the vanilla InterGen baseline (all metrics at their worst), adding $\mathcal{L}_{\text{root}}$ reduces positional error to 0.0. Adding $\mathcal{L}_{\text{overlap}}$ eliminates overlap to 0.0. Adding $\mathcal{L}_{\text{region}}$ reduces region violations from 0.958 to 0.043. Finally, adding $\mathcal{L}_{\text{orient}}$ brings orientation error from 1.0 down to 0.208—the residual error reflects the inherent ambiguity of orientation in complex motions. Notably, ==each penalty targets its corresponding metric without significantly degrading others, validating the modular design of our loss function.==

<span style="color:red">GT shows individual penalties degrade other metrics; improvements occur only when combined, not per-penalty.</span>

## 5.6. Runtime Analysis

Noise optimization introduces additional inference time relative to vanilla diffusion sampling. On a single NVIDIA A100 GPU, generating a two-person sequence of $T = 150$ frames with PINO requires approximately 90 seconds, compared to under 5 seconds for standard InterGen. For a five-person group, the total generation time scales approximately linearly with group size, reaching roughly 6 minutes. Although longer than baseline sampling, this overhead avoids any retraining—a process requiring hours on the full InterHuman dataset. The per-person optimization steps are independent given the pivot motion and can therefore be parallelized across GPUs to reduce wall-clock time further.

## 5.7. Discussion

Taken together, the experimental results reveal a consistent theme: PINO's training-free noise optimization effectively decouples physical plausibility from semantic quality. Unlike approaches that enforce physical constraints through auxiliary loss terms during training—which risk overfitting

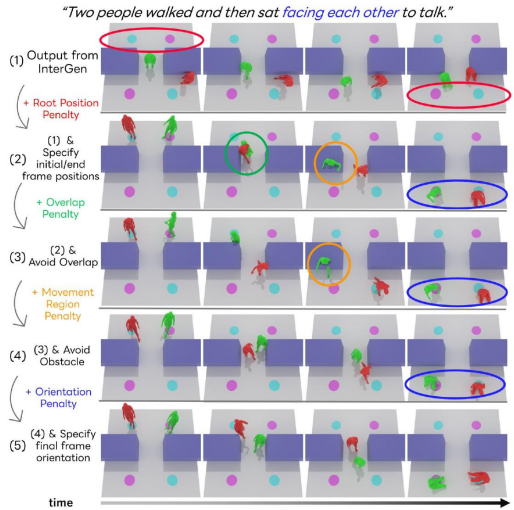

*"Two people walked and then sat facing each other to talk."*

(1) Output from InterGen

+ Root Position Penalty

(1) & Specify initial/end frame positions

+ Overlap Penalty

(2) & Avoid Overlap

+ Movement Region Penalty

(3) & Avoid Obstacle

+ Orientation Penalty

(4) & Specify final frame orientation

time

Figure 4. Visualization of the ablation study. Left to right: InterGen baseline, adding $\mathcal{L}_{root}$, $+\mathcal{L}_{overlap}$, $+\mathcal{L}_{region}$, $+\mathcal{L}_{orient}$. Each penalty progressively enforces its corresponding spatial constraint while preserving overall motion quality.

the model to specific physics regimes—our inference-time optimization adapts to any spatio-temporal constraint expressible as a differentiable penalty. This flexibility makes PINO particularly well-suited for applications requiring on-the-fly customization, such as interactive animation tools or scenario-specific motion synthesis for robotics simulation. Moreover, the ablation results in Section 5.5 confirm that each penalty term targets its respective artifact independently, which simplifies hyperparameter tuning: practitioners can selectively enable or disable individual constraints without destabilizing the overall optimization. The consistent gains in physical plausibility metrics across two-person, multi-person, and long-duration settings further validate the generality of the pivot-anchored pairwise decomposition as a scalable interaction generation paradigm.

One notable observation is that noise optimization acts as a form of post-hoc constraint satisfaction rather than a learned prior; consequently, the quality of generated motions is bounded by the capacity of the underlying InterGen model. When a text prompt specifies an interaction underrepresented in InterGen's training distribution, the optimizer may require additional iterations to balance the semantic and physical objectives simultaneously. A promising direction for future work is to combine our inference-time optimization with lightweight fine-tuning on in-domain data, achieving improved semantic fidelity at lower computational cost. Additionally, the pairwise generation paradigm assumes group interactions can be

adequately captured through bilateral relationships with a shared pivot. While this holds for common scenarios—group conversations, coordinated exercises, or team handshakes—it may be less appropriate for interactions requiring simultaneous coupling among three or more persons, such as a group lift, where higher-order dependencies cannot be cleanly decomposed into pairwise terms.

It is also worth highlighting that PINO's modular penalty design provides a natural interface for domain experts to specify interaction semantics. A choreographer, for instance, could define custom region and orientation constraints to stage a group dance performance, while a robotics engineer could specify tight spacing and collision-avoidance distances for coordinated manipulation tasks. This versatility positions noise optimization as a general-purpose inference-time control mechanism for generative motion models, extending well beyond the academic benchmarks considered here.

## 6. Limitations

While PINO achieves strong results, several limitations remain. First, the overlap penalty operates on root (pelvis) distances and does not model the full 3D body mesh; in highly contorted poses, slight limb penetrations can still occur. Second, noise optimization increases inference time compared to vanilla diffusion sampling, as gradients must be backpropagated through the entire denoising chain at each optimization iteration. Third, as the group size grows, the pivot-anchored approach may limit the diversity of person $k$'s motion, since it is always conditioned on person 1. Exploring alternative graph topologies (e.g., ring or tree structures) for pivot assignment is left as future work. Finally, the current framework generates all persons' motions in a fixed XZ ground plane, which may not capture all interaction geometries (e.g., stacked, airborne motions). Furthermore, backpropagating gradients through the full DDIM denoising chain requires retaining the entire computation graph in GPU memory, which can become a bottleneck for very long sequences ($T \gg 150$) or large groups on memory-constrained hardware; gradient checkpointing strategies could mitigate this cost in future implementations.

## 7. Conclusion

We have presented PINO, a training-free framework for generating physically plausible multi-person interactions of arbitrary group size. By decomposing group dynamics into a sequence of pairwise diffusion generations anchored by a pivot character, and by optimizing the initial noise with differentiable physics-based penalties, PINO achieves near-zero body overlap and dramatically reduced penetration volumes without any additional training. Per-

pair text prompt switching enables fine-grained role assignment within large groups. A motion inpainting strategy with boundary smoothness penalties extends the framework to long-duration sequences. Extensive experiments on Inter-Human demonstrate consistent improvements in physical plausibility metrics while maintaining competitive semantic quality.

# PINO: Person-Interaction Noise Optimization for Long-Duration and Customizable Motion Generation of Arbitrary-Sized Groups

Anonymous ICCV Submission

## Abstract

*Generating text-driven human interactions beyond two people remains difficult because interaction complexity grows combinatorially with group size, while existing models are typically trained for a fixed number of actors and often produce collisions, shared-prompt ambiguity, or limited spatial control. We present PINO, a training-free framework that composes arbitrary-sized group interactions from a pretrained two-person diffusion model by optimizing the initial diffusion noise at inference time. PINO adds one new character at a time around a shared pivot person, reusing the pretrained model through masked conditioning so previously generated motion is preserved while pair-specific prompts define the role of the newly added participant. The optimization objective combines collision-aware and spatio-temporal penalties, enabling overlap avoidance, root-position control, region constraints, and orientation control without retraining the backbone model. PINO also extends interactions to longer durations through inpainting-based continuation with optimized unobserved segments. When instantiated on InterGen, PINO reduces overlap from 0.119 to 0.000 and penetration volume from 3112.72 to 275.65 in two-person generation while preserving semantic quality. In multi-person generation, it reduces overlap from 0.766 to 0.000 and penetration volume from 25215.78 to 785.01, and it continues to scale favorably up to five people. For motion extension, PINO improves both semantic fidelity and physical plausibility, reducing FID from 20.390 to 18.930 and overlap from 0.089 to 0.000. These results show that inference-time noise optimization is a practical route to controllable and physically coherent group interaction synthesis without additional training.*

## 1. Introduction

Text-driven human motion generation has advanced rapidly with diffusion models for single-person motion and two-person interaction synthesis [2, 6, 11, 14]. However, generating realistic interactions for arbitrary-sized groups remains substantially harder than the two-person case. The main difficulty is not only modeling additional bodies, but

preserving coherent semantics while preventing trivial failure modes such as body overlap, interpenetration, and role ambiguity across multiple participants.

Existing approaches address this challenge from two main directions. One line trains dedicated multi-person generators that directly model several people at once [1, 9]. This offers expressive generation, but the architecture and training data are tied to the group distributions seen during training. Another line keeps pretrained priors fixed and applies inference-time control or optimization [5, 7, 12]. These methods improve flexibility, but they do not directly exploit the structure of human group interactions as a composition of localized pairwise relations.

Our starting observation is that many group interactions can be organized around a shared pivot person. For example, a central character may greet one person, guide another, and coordinate with several surrounding partners. This suggests a simple decomposition: instead of asking one model invocation to explain the entire group, we can sequentially compose pairwise interactions between the pivot and each newly added person. A pretrained two-person interaction diffusion model already contains a strong prior over pose realism, timing, and semantic interaction. The missing component is a mechanism that preserves this prior while enforcing group-level physical consistency and user control.

Based on this idea, we propose **PINO** (**P**erson-**I**nteraction **N**oise **O**ptimization), a training-free framework for long-duration and customizable group interaction generation. PINO uses a pretrained two-person diffusion model as a frozen backbone, then optimizes the initial diffusion noise for each pairwise generation step. Masked conditioning keeps the pivot person fixed, while the newly added character is synthesized from a pair-specific text prompt. The optimization objective is defined directly in motion space and includes penalties for overlap avoidance, start/end root positions, movement regions, and orientation. This retains the semantic prior of the pretrained model while giving the user precise inference-time control.

PINO also supports long-duration interaction generation. Rather than retraining a dedicated long-horizon model, we extend a motion sequence by treating known frames as fixed conditions and optimizing the unobserved segment

1

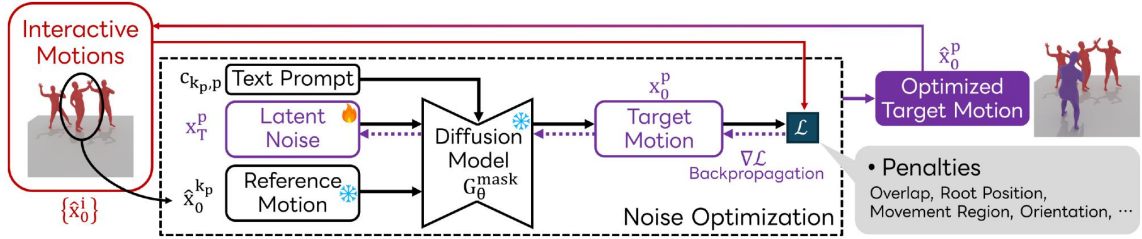

Figure 1. **Person-Interaction Noise Optimization (PINO).** Starting from a pretrained two-person diffusion model, PINO generates an initial pair and then sequentially adds new participants around a pivot person through masked conditioning. The initial noise for the new pair is optimized with physics-aware and spatio-temporal penalties, enabling collision avoidance and controllable group composition without retraining.

through inpainting. This preserves continuity while keeping the same inference-time control mechanism used for group composition.

Experiments on InterGen-based generation show a consistent pattern: PINO yields the largest gains in physical plausibility and scalability, while preserving or modestly improving semantic metrics. In two-person generation, PINO eliminates overlap and sharply reduces penetration volume without degrading FID. In multi-person generation, the gap becomes larger, especially as the number of people increases. In motion extension, PINO improves both semantic fidelity and non-semantic realism.

Our contributions are threefold. First, we introduce a training-free framework that composes arbitrary-sized group interactions from a pretrained two-person diffusion model through sequential pivot-based generation. Second, we formulate inference-time person-interaction noise optimization with collision-aware and spatio-temporal penalties, enabling customizable control without retraining. Third, we show that the same mechanism extends to long-duration interaction generation through inpainting-based continuation.

## 2. Related Work

### 2.1. Diffusion Models for Human Motion

Diffusion models have become a standard backbone for text-to-motion generation because they produce diverse and realistic trajectories while supporting flexible conditioning [3, 10, 11, 14]. Subsequent work improved fidelity and controllability through masked modeling, latent-space generation, or stronger priors [2, 8]. These methods focus primarily on single-person motion and do not directly address multi-person physical constraints. PINO instead treats a pretrained motion diffusion model as a reusable prior and moves the control problem to inference time.

### 2.2. Interactive and Multi-Person Motion Generation

Two-person interaction generation has recently become feasible with specialized architectures and datasets, especially through InterGen [6]. More recent interactive generators further improve fidelity and reaction modeling, but they still target a fixed number of participants during training. For larger groups, FreeMotion [1] and Shan et al. [9] move toward open-domain multi-person generation, yet they model multiple people jointly and therefore inherit the difficulty of scaling network design and training data to larger groups. PINO takes a different route: it reuses a pretrained two-person model and composes larger groups sequentially, which makes the number of generated people an inference-time choice rather than a training-time design decision.

### 2.3. Inference-Time Control and Optimization

Inference-time control has emerged as an attractive alternative to retraining motion generators. Guided Motion Diffusion [4] and OmniControl [13] incorporate spatial constraints for single-person motion generation. Programmable Motion Generation [7] and Diffusion Noise Optimization [5] show that optimizing latent variables or diffusion noise can satisfy diverse motion-space constraints without task-specific retraining. In the multi-person setting, InterControl [12] uses joint-level interaction control for zero-shot synthesis. PINO is closest in spirit to these training-free methods, but differs in two ways: it explicitly decomposes group generation into pivot-centered pairwise interactions, and it uses the optimization to preserve physical consistency among already generated participants while allowing prompt switching across pairs.

## 3. Method

### 3.1. Problem Setup

We assume access to a pretrained text-conditioned two-person diffusion model $G$ such as InterGen [6]. Given a text

| Method | #Person in Network Design | #Person in Training Data | Order of Multi-Person Generation | Method |
|---|---|---|---|---|
| InterControl [12] | One | One | Sequential | ControlNet |
| FreeMotion [1] | One | Two | Sequential | ControlNet |
| Shan et al. [9] | Multiple | Multiple | Concurrent | Supervised Training |
| ProgMoGen [7] | One | Training-free | N/A (Single-Person) | Noise Optimization |
| Ours | Two (or multiple) | Training-free | Sequential | Noise Optimization |

Table 1. Comparison of recent multi-person interaction generation methods, including single-person method ProgMoGen [7].

prompt $p$, the model maps an initial Gaussian noise tensor $z$ to a motion sequence $\hat{X} = G(z, p)$ containing two interacting people. Each person is represented by a sequence of motion features; in the InterGen instantiation, a single person uses 262 features per frame, and optimization is performed on 210-frame windows.

Our goal is to generate a group of $N$ interacting people, potentially with pair-specific prompts and motion constraints, without changing the pretrained weights of $G$. We also want the same mechanism to support motion extension beyond the native window length of the pretrained model.

### 3.2. Pivot-Based Pairwise Composition

The key design choice in PINO is to organize group generation around a *pivot* person shared across pairwise generations. We first sample a two-person interaction from prompt $p_{1,2}$ to obtain the pivot motion $X^{(1)}$ and the second participant motion $X^{(2)}$. To add a new participant $k$, we invoke the same two-person diffusion model again with a new prompt $p_{1,k}$, but now one input slot is fixed to the pivot motion:

$$\hat{X}^{(1,k)} = G(z_k, p_{1,k}; X^{(1)}, M), \tag{1}$$

where $M$ is a binary mask that preserves the pivot features and leaves the new participant unconstrained.

This masked conditioning is implemented through the inpainting interface of the frozen backbone. Concretely, the known coordinates of the pivot motion are copied into the denoiser output at every reverse step, so optimization only modifies the unconstrained participant. The result is a new motion $X^{(k)}$ that interacts with the pivot while remaining compatible with previously generated people. Repeating this process sequentially yields groups of arbitrary size.

This decomposition has two benefits. First, it avoids asking a single prompt to explain the roles of all participants at once. Each pair can have its own prompt, which allows prompt switching and clearer role assignment. Second, it leverages a pretrained interaction prior that already models close-range coordination, while group-level consistency is handled by optimization. The pivot assumption is also

computationally useful. Keeping one actor fixed across generations propagates a stable temporal reference into every new pair, which reduces the number of unconstrained bodies that the optimizer must coordinate simultaneously. In practice, this turns a hard many-body search problem into a sequence of smaller constrained updates while still allowing non-pivot participants to interact indirectly through the shared spatial layout.

### 3.3. Person-Interaction Noise Optimization

For each generation step, PINO optimizes the initial diffusion noise rather than retraining or fine-tuning the backbone. Let $z_k^0$ be an initial random noise sample for the pair consisting of the pivot and a new target person. PINO solves

$$z_k^\star = \arg\min_{z_k} \ \lambda_{ov}\mathcal{L}_{\text{overlap}} + \lambda_{\text{root}}\mathcal{L}_{\text{root}} + \lambda_{\text{reg}}\mathcal{L}_{\text{region}} + \lambda_{\text{ori}}\mathcal{L}_{\text{orientation}} \tag{2}$$

where all losses are computed on the decoded motion.

**Overlap loss.** The default objective encourages the newly generated person to maintain a valid distance from the pivot and from all previously generated non-pivot participants. In the released implementation, this is expressed through root-distance penalties on the pelvis trajectory, with a minimum distance threshold for collision avoidance and an optional maximum threshold to prevent degenerate separations. When generating the $k$-th person, the loss also pushes that person away from already generated people other than the pivot.

**Root-position loss.** To support temporal anchors, PINO can constrain the start and end root positions of one or both participants. This is useful when the user needs a character to enter or exit at a prescribed location, or when the motion must align with external timing constraints.

**Region loss.** PINO can restrict a participant to remain inside or outside a spatial region. The implementation supports circular or rectangular regions on the ground plane, enabling scene-aware placement or motion zoning without retraining a scene-conditioned model.

**Orientation loss.** An orientation penalty aligns the participant's facing direction with a target direction. This is especially useful for prompts whose semantics depend on mutual attention, such as facing a pivot before approaching or gesturing.

The optimization is fully differentiable because gradients propagate from the motion-space objective through the denoising process back to the initial noise. In practice, we use Adam for up to 100 iterations with early stopping, following the optimization loop implemented in the released code.

3

**Algorithm 1** PINO inference for group composition

---
1: Generate $(X^{(1)}, X^{(2)}) \leftarrow G(z_2, p_{1,2})$
2: **for** $k = 3$ to $N$ **do**
3:    Sample $z_k$ and fix pivot motion $X^{(1)}$ with mask $M$
4:    Optimize $z_k$ using Eq. (2)
5:    Decode $X^{(k)} \leftarrow G(z_k^\star, p_{1,k}; X^{(1)}, M)$
6: **end for**
7: **return** $\{X^{(i)}\}_{i=1}^N$

---

### 3.4. Prompt Switching and Customizable Roles

Because each pairwise generation step uses its own text prompt, PINO naturally supports prompt switching across participants. A single group motion can therefore mix semantically different relations around the same pivot, such as handshaking with one person while addressing or guiding another. This is difficult for shared-prompt approaches, which often blur individual roles when several people must simultaneously satisfy one global textual description.

### 3.5. Long-Duration Motion Extension

The same formulation can extend interactions beyond the native duration of the pretrained model. We keep a subset of frames fixed as observed motion, mask them through the inpainting interface, and optimize the noise for the remaining interval. This reuses the pretrained prior for local motion realism while allowing the unobserved portion to adapt to collision-aware and control-oriented objectives. In practice, the known segment acts as a temporal anchor at the transition boundary, producing longer interactions without training a separate long-horizon model.

## 4. Experiments

### 4.1. Experimental Setup

We instantiate PINO on InterGen [6] and denote the resulting system as *PINO-InterGen*. The evaluation follows the InterHuman protocol used by the provided codebase and reports both semantic and non-semantic metrics. Semantic quality is measured by FID, diversity, and R-precision when available. Physical plausibility is measured by overlap, penetration volume, foot skating, and maximum acceleration. For multi-person generation, we compare against InterGen with inpainting-based composition and against FreeMotion [1] where available.

Unless otherwise noted, PINO keeps the pretrained generator frozen and applies optimization only at inference time. This matters because improvements cannot be attributed to additional supervised training or changes in backbone capacity.

| Method | Overlap↓ | PenVol. (cm³) ↓ | Foot Skate ↓ | Max Acc. ↓ |
|---|---|---|---|---|
| GT | 0.029 | 471.75 | 0.099 | 0.074 |
| InterGen | 0.119 | 3112.72 | **0.124** | **0.034** |
| PINO-InterGen | **0.000** | **275.65** | 0.125 | **0.034** |

(a) Non-semantic metrics.

| Method | FID ↓ | Div. → | R-Prec. ↑ (Top3) |
|---|---|---|---|
| GT | 0.983 | 7.921 | 0.715 |
| InterGen | 13.278 | 7.793 | 0.674 |
| PINO-InterGen | **13.163** | **7.904** | **0.675** |

(b) Semantic metrics.

Table 2. Evaluation of two-person interaction generation while avoiding overlap.

| Method | Overlap↓ | PenVol.↓ | Foot Skate↓ | Max Acc.↓ |
|---|---|---|---|---|
| InterGen (inpainting) | 0.766 | 25215.78 | 0.099 | **0.030** |
| PINO-InterGen | **0.000** | **785.01** | **0.091** | 0.031 |

Table 3. Evaluation of multi-person interaction generation

### 4.2. Two-Person Interaction Generation with Overlap Avoidance

We first evaluate whether PINO can improve physical plausibility in the base two-person setting without sacrificing semantic quality. Table 2 shows that it does. PINO-InterGen reduces overlap from 0.119 to 0.000 and penetration volume from 3112.72 to 275.65. At the same time, FID slightly improves from 13.278 to 13.163, diversity increases from 7.793 to 7.904, and R-precision is marginally higher.

These results are important because they isolate the effect of noise optimization from the more complex multi-person setting. The gain is not coming from a different motion prior; it comes from using the same pretrained prior more effectively under a physically informed objective.

### 4.3. Multi-Person Interaction Generation

We next evaluate the main setting of interest: composing larger groups by sequentially adding participants around a pivot. Table 3 compares PINO-InterGen against InterGen with inpainting-based multi-person composition. PINO reduces overlap from 0.766 to 0.000 and penetration volume from 25215.78 to 785.01, while also slightly improving foot skating. Maximum acceleration is nearly unchanged (0.031 vs. 0.030), indicating that the gain does not come from overly damped motions.

To study scaling, Table 4 reports pairwise quality as the group grows from two to five people using person 1 as the pivot. PINO-InterGen consistently achieves the best FID for every pair and dramatically lowers overlap at each step. The advantage becomes larger as more participants are added: for pair $(1, 5)$, FID drops from 19.501 to 16.911 and overlap drops from 0.977 to 0.069 compared with InterGen. Earlier pairs remain near zero overlap, which indicates that sequen-

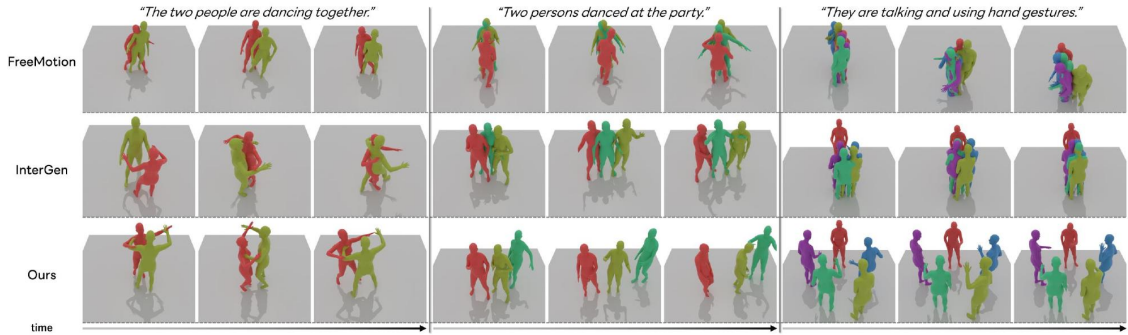

Figure 2. Comparison of the generated multi-person interaction. The visualization highlights that PINO produces cleaner spacing and fewer severe collisions when new participants are added sequentially around a pivot.

| Method | Pair | FID↓ | Overlap↓ | Foot Skate↓ | Max Acc.↓ |
|--------|------|------|----------|-------------|-----------|
| FreeMotion | (1,2) | 19.383 | 0.255 | **0.112** | **0.034** |
| | (1,3) | 25.184 | 0.806 | **0.122** | 0.038 |
| | (1,4) | 25.812 | 0.964 | 0.127 | 0.040 |
| | (1,5) | 25.671 | 0.991 | 0.130 | 0.042 |
| InterGen | (1,2) | 13.100 | 0.119 | 0.126 | **0.034** |
| | (1,3) | 19.414 | 0.799 | 0.125 | **0.036** |
| | (1,4) | 19.386 | 0.937 | **0.124** | **0.038** |
| | (1,5) | 19.501 | 0.977 | **0.124** | 0.039 |
| PINO-InterGen | (1,2) | **12.920** | **0.000** | 0.127 | **0.034** |
| | (1,3) | **17.170** | **0.002** | 0.135 | 0.043 |
| | (1,4) | **16.890** | **0.019** | 0.138 | 0.043 |
| | (1,5) | **16.911** | **0.069** | 0.142 | 0.045 |

Table 4. Evaluation of multi-person interaction generation. From pair (1,2), interactions are generated incrementally using 1 as a pivot. Overlap is among all individuals generated up to that step.

tial composition does not accumulate severe geometric failures as quickly as the baseline.

### 4.4. Long-Duration Motion Extension

We finally evaluate motion extension. In the semantic evaluation of Table 5, PINO-InterGen improves FID from 20.390 to 18.930, reduces overlap from 0.089 to 0.000, lowers foot skating from 0.081 to 0.054, and substantially lowers maximum acceleration from 0.036 to 0.025. In the non-semantic extension benchmark of Table 6, PINO likewise improves every reported metric, including overlap (0.672 to 0.031), penetration volume (15622.16 to 1375.24), and foot skating (0.070 to 0.045).

Together, these results suggest that inference-time optimization is effective not only for adding more people, but also for extending interaction duration while preserving realism at the transition.

| Method | FID↓ | Overlap↓ | Foot Skate↓ | Max Acc.↓ |
|--------|------|----------|-------------|-----------|
| InterGen | 20.390 | 0.089 | 0.081 | 0.036 |
| PINO-InterGen | **18.930** | **0.0** | **0.054** | **0.025** |

Table 5. Evaluation of semantic elements in two-person motion extension. Extended motions are cropped to match the ground truth length for evaluation.

| Method | Overlap↓ | PenVol.↓ | Foot Skate↓ | Max Acc.↓ |
|--------|----------|----------|-------------|-----------|
| InterGen (inpainting) | 0.672 | 15622.16 | 0.070 | 0.044 |
| PINO-InterGen | **0.031** | **1375.24** | **0.045** | **0.043** |

Table 6. Evaluation of non-semantic elements in motion extension.

## 5. Analysis

### 5.1. Ablation on Motion Penalties

Table 7 isolates the role of different inference-time penalties. Starting from plain InterGen, adding the root-position loss eliminates positional error but does not address collisions and substantially increases region violations. Adding the overlap loss removes overlap entirely, confirming that the main physical gain in PINO is driven by explicit geometric control rather than being an incidental byproduct of prompt decomposition. Adding the region loss produces the best region-violation score (0.043), while the orientation loss sharply reduces orientation error from 1.0 to 0.208.

This ablation supports the design of Eq. (2): no single penalty solves all control requirements. Instead, the strength of PINO is that these constraints can be combined as needed at inference time, while the pretrained motion prior remains fixed.

### 5.2. Qualitative Observations

Fig. 2 and Fig. 4 provide qualitative evidence consistent with the metrics. Compared with direct multi-person com-

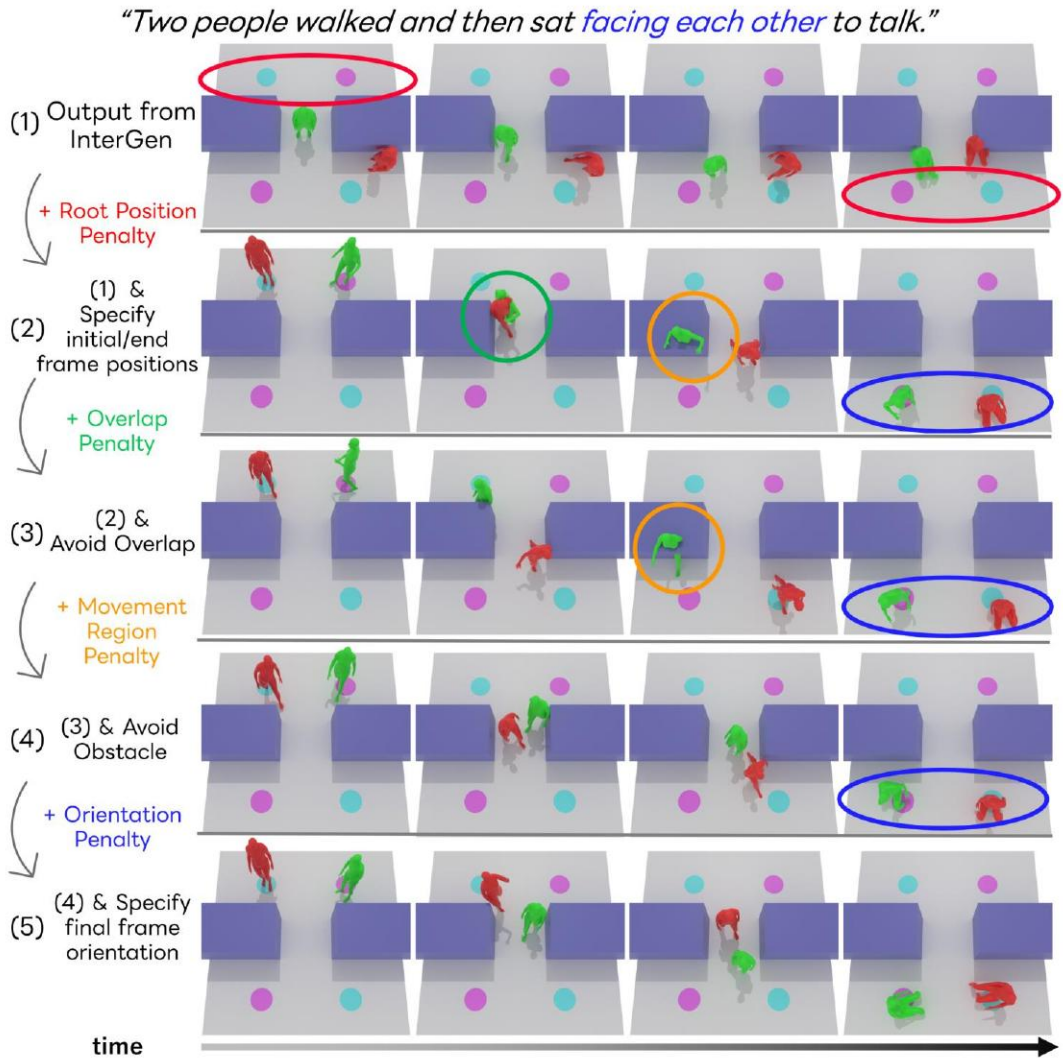

Figure 3. Visualization of the ablation study. Different penalties affect different aspects of the generated motion, with overlap-oriented terms primarily improving physical plausibility and orientation/region terms improving controllability.

position, PINO maintains clearer personal space and produces less entangled body configurations. The visualizations also suggest why pivot-based prompt switching is useful: the new participant can adopt a distinct role relative to the pivot instead of collapsing toward an averaged group behavior.

### 5.3. Discussion

The empirical pattern across all benchmarks is consistent. PINO's largest gains appear in collision-related metrics, especially overlap and penetration volume, and the margin widens as the composition problem becomes harder. Semantic fidelity is preserved in the base two-person setting and improves in motion extension, which indicates that optimizing the initial noise does not simply trade semantics

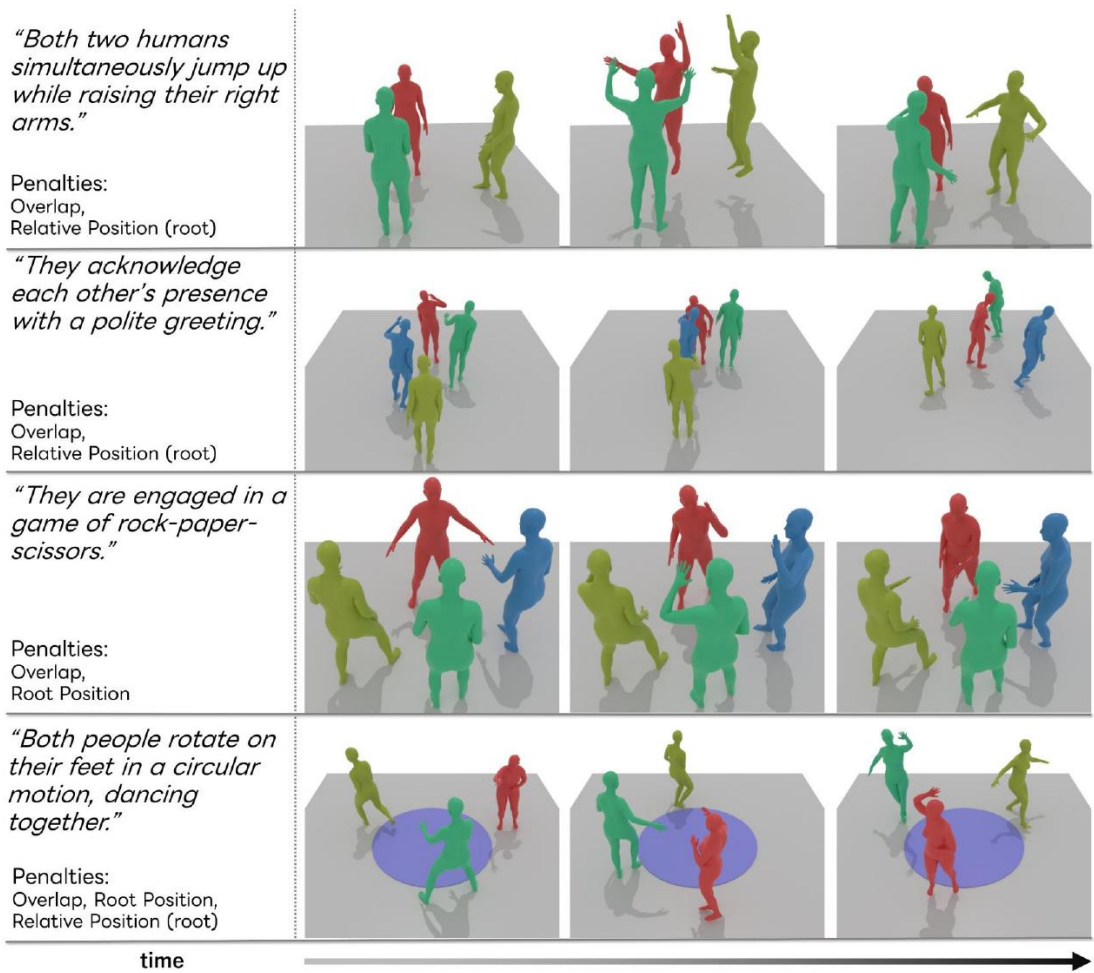

*"Both two humans simultaneously jump up while raising their right arms."*

Penalties:
Overlap,
Relative Position (root)

*"They acknowledge each other's presence with a polite greeting."*

Penalties:
Overlap,
Relative Position (root)

*"They are engaged in a game of rock-paper-scissors."*

Penalties:
Overlap,
Root Position

*"Both people rotate on their feet in a circular motion, dancing together."*

Penalties:
Overlap, Root Position,
Relative Position (root)

time

Figure 4. Qualitative results of multi-person interaction generation. PINO supports customizable pairwise interactions within a larger group while maintaining physically cleaner spacing.

for geometry. At the same time, some motion smoothness metrics remain close to the baseline rather than improving dramatically, suggesting that PINO's main benefit is better physical feasibility and controllability rather than a wholesale change in motion style.

## 6. Conclusion

We presented PINO, a training-free framework for arbitrary-sized and long-duration human interaction generation. PINO composes a group around a pivot person using a pretrained two-person diffusion model, and it optimizes the initial diffusion noise with motion-space penalties to en-

force physical plausibility and user control. Experiments show that this simple decomposition scales better than direct composition baselines, substantially reduces overlap and penetration, and supports motion extension without retraining. These results indicate that inference-time noise optimization is a strong and practical interface for turning pretrained interaction models into controllable group motion generators.

| Method | Pos. Err.↓ | Overlap↓ | Reg. Viol.↓ | Orient. Err.↓ |
|---|---|---|---|---|
| InterGen | 1.0 | 0.292 | 0.500 | 1.0 |
| + $\mathcal{L}_{\text{root}}$ | **0.0** | 0.333 | 0.917 | 1.0 |
| + $\mathcal{L}_{\text{overlap}}$ | **0.0** | **0.0** | 0.958 | 1.0 |
| + $\mathcal{L}_{\text{region}}$ | 0.083 | **0.0** | **0.043** | 1.0 |
| + $\mathcal{L}_{\text{orientation}}$ | 0.083 | 0.043 | 0.083 | **0.208** |

Table 7. Ablation study on motion penalties. Metrics include positional errors (Pos. Err.↓), overlap (Overlap↓), region violations (Reg. Viol.↓), and orientation errors (Orient. Err.↓). Positional errors are measured with a threshold of 20 cm, region violations with a threshold of 10 cm, and orientation errors with a threshold of 20 degrees. Lower values indicate better performance.


\let\clearpage\relax

\end{document}